%% file: main.tex
\documentclass[manuscript,screen,nonacm]{acmart}
\makeatletter
\let\@authorsaddresses\@empty
\makeatother

\renewcommand\footnotetextcopyrightpermission[1]{}
\usepackage{url}
\usepackage{listings}

\usepackage[most]{tcolorbox}
\usepackage{xcolor}

\definecolor{reviewerpink}{RGB}{248,228,233}

\newtcbox{\rb}{
  on line,
  boxrule=0pt,
  colback=reviewerpink,
  colframe=reviewerpink,
  arc=3pt,
  left=2pt,
  right=2pt,
  top=2pt,
  bottom=2pt,
  boxsep=0pt
}

\usepackage{subcaption}
\usepackage{cleveref}
\usepackage{algorithm}
\usepackage{algorithmic}
\usepackage{multirow}
\begin{document}
\title{Your Recourse, My Loss? Algorithmic Recourse under Shared  Constraints}

\author{Zahra Khotanlou}
\affiliation{%
  \institution{Electrical and Computer Engineering Department, University of Waterloo}
  \city{Waterloo}
  \country{Canada}}
\email{zkhotanl@uwaterloo.ca}

\author{Kate Larson}
\affiliation{%
  \institution{Cheriton School of Computer Science, University of Waterloo}
  \city{Waterloo}
  \country{Canada}
}
\email{kate.larson@uwaterloo.ca}

\author{Amir-Hossein Karimi}
\affiliation{%
 \institution{Electrical and Computer Engineering Department, University of Waterloo}
 \city{Waterloo}
 \country{Canada}}
\email{amirh.karimi@uwaterloo.ca} 

\renewcommand{\shortauthors}{Khotanlou et al.}


\begin{abstract}
  \input{sections/Abstract}
\end{abstract}





\maketitle

\input{sections/Introduction}
\input{sections/Many-to-Many_Recourse_Optimization}
\input{sections/Minimize_Welfare_Gap}

\input{sections/Fairness}

\input{sections/Experiments}
\input{sections/Discussion}
\input{sections/Conclusion}







\bibliographystyle{ACM-Reference-Format}
\bibliography{references}

\newpage
\appendix
\section*{Appendices}
\input{sections/Appendix}
\end{document}

%% file: sections/Abstract.tex
Decision makers are increasingly relying on machine learning in sensitive situations.
In such settings, \emph{algorithmic recourse} aims to provide individuals with actionable and minimally costly steps to reverse unfavorable AI-driven decisions.
While existing research predominantly focuses on single-individual (i.e., \emph{seeker}) and single-model (i.e., \emph{provider}) scenarios, real-world applications often involve multiple interacting stakeholders.
Optimizing outcomes for seekers under an \emph{individual welfare} approach overlooks the inherently multi-agent nature of real-world systems, where individuals interact and compete for limited resources.
Accordingly, we extend algorithmic 
recourse to a many-to-many setting with capacity constraints, where individually computed recourse recommendations no longer compose independently and stakeholder interactions directly affect recourse validity.
To address this, we introduce a novel framework for multi-agent algorithmic recourse that accounts for multiple recourse seekers and recourse providers.
We model this many-to-many interaction as a \emph{capacitated weighted bipartite matching problem}, where matches are guided by both recourse cost and provider capacity.
Edge weights, reflecting recourse costs, are optimized for \emph{social welfare} while quantifying the welfare gap between individual welfare and this collectively feasible outcome.
We propose a three-layer optimization framework: (1) basic capacitated matching, (2) optimal capacity redistribution to minimize the welfare gap, and (3) cost-aware optimization balancing welfare maximization with capacity adjustment costs.
We further extend our approach to support inequality-averse system objectives through a concave social-welfare formulation that prioritizes the most disadvantaged seekers.
Experimental validation on synthetic and real-world datasets demonstrates that our framework enables the many-to-many algorithmic recourse to achieve near-optimal welfare with minimum modification in system settings.
Our results also show an inherent efficiency–fairness trade-off, demonstrating how recourse systems can be designed to balance aggregate welfare with distributive considerations in resource-constrained environments.
This work extends algorithmic recourse from individual recommendations to system-level design, providing a tractable path toward higher social welfare while maintaining individual actionability.
%

%% file: sections/Introduction.tex
\section{Introduction}
\label{sec:intro}
\begin{figure}[t]
  \centering
  \Description{Matching Settings}
  \begin{subfigure}[c]{0.32\linewidth}
    \centering
    \includegraphics[width=\linewidth]{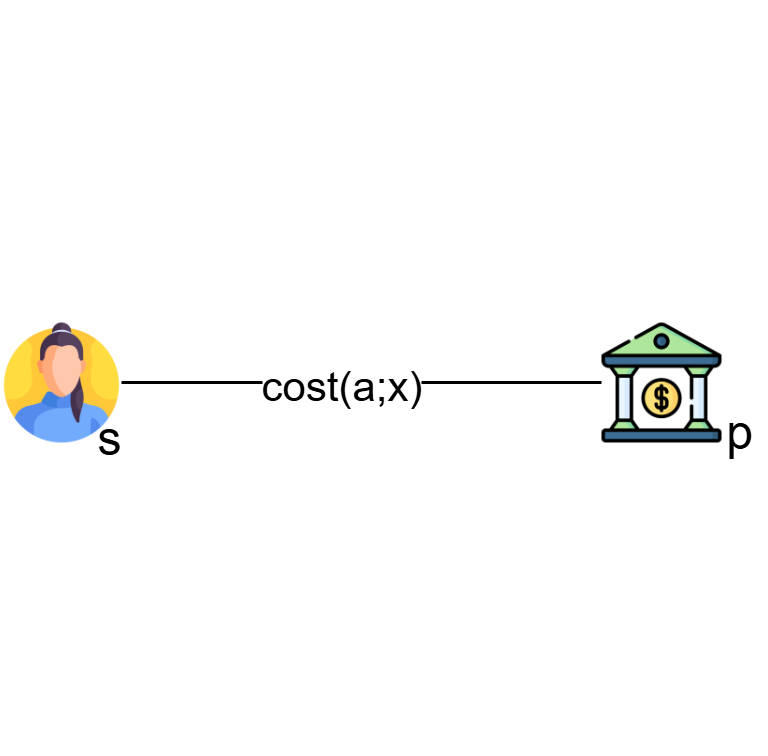}
    \caption{One-to-one Recourse Matching}
    \label{fig:one-to-one}
  \end{subfigure}
  \hfill
  \begin{subfigure}[c]{0.32\linewidth}
    \centering
    \includegraphics[width=\linewidth]{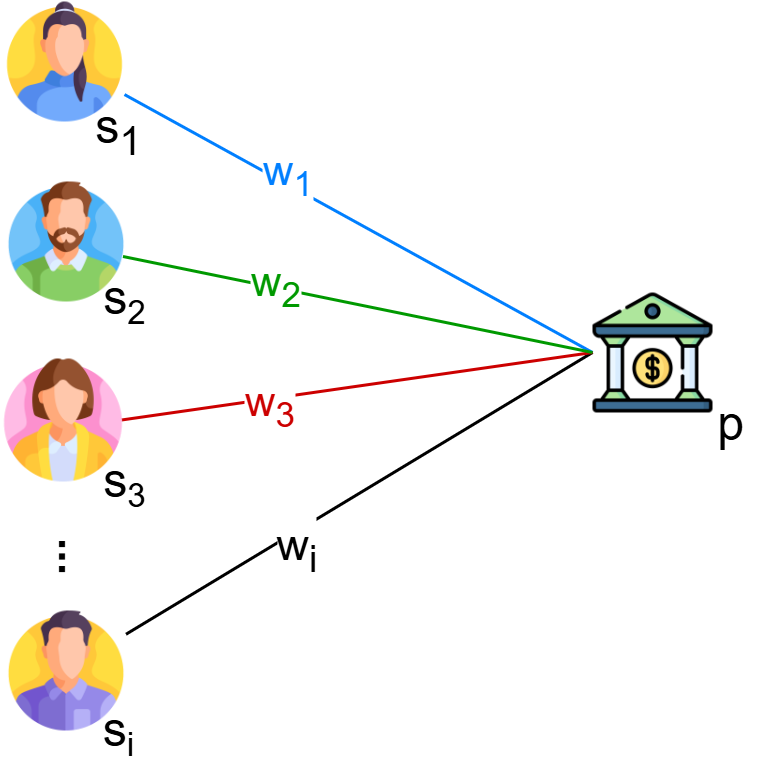}
    \caption{Many-to-one Setting}
    \label{fig:many-to-one}
  \end{subfigure}
  \hfill   
  \begin{subfigure}[c]{0.32\linewidth}
    \centering
    \includegraphics[width=\linewidth]{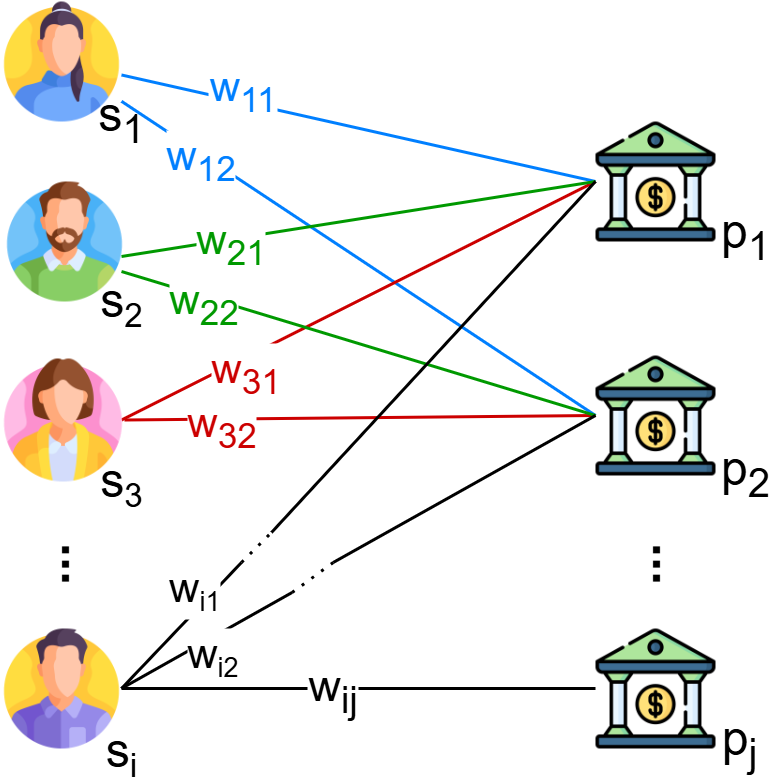}
    \caption{Many-to-many Recourse Matching}
    \label{fig:many-to-many}
  \end{subfigure}
  \caption{Various algorithmic recourse setups: (a) the classical one-to-one setting, where an individual \(s\) seeks recourse recommendations from a provider \(p\) with minimal cost required to reverse the output; (b) the many-to-one setting, where multiple individuals are seeking recourse from a single recourse provider. (c) Our proposed many-to-many framework generalizes prior settings by simultaneously optimizing for multiple recourse seekers and providers.}
  \label{fig:matching-settings}
\end{figure}

AI decision-making systems rapidly apply predictive models to support individuals in various contexts, e.g., loan approvals, medical treatments, or bail decisions~\citep{voigt_eu_2017}. 
The increasing reliance of humans on these algorithmic decision-making systems and their significant impact on areas such as finance, healthcare, and criminal justice have raised concerns regarding the transparency and fairness of these automated systems. 
Driven by AI policy regulations and the idea of a "right to explanation," algorithmic recourse is an emerging field that aims to provide individuals affected by negative, high-stakes algorithmic decisions with recommendations on how to reverse those outcomes~\citep{gdpr_regulation_2016, verma_counterfactual_2024}.
Therefore, algorithmic recourse refers to the systematic process of reversing unfavorable decisions made by algorithms across various counterfactual scenarios~\citep{wachter_counterfactual_2018}. 
It encompasses the necessary actions individuals must take to achieve a favorable outcome, serving as a foundation for temporally extended agency and trust in automated decision-making systems~\citep{karimi_algorithmic_2021}. %
This concept is essential to ensure automated decisions are understandable and to enable individuals to engage with and contest these decisions~\citep{doshi-velez_towards_2017, gunning_darpas_2019}.
Existing studies on algorithmic recourse predominantly address how the individual would need to change their attributes to achieve the desired outcome~\citep{karimi_survey_2022}. 
Such settings generally assume a single individual impacted by a single decision-making model as shown in~\Cref{fig:one-to-one}.
In real-world scenarios, however, AI decision-making systems (i.e., \emph{providers}) often interact with multiple individuals whose actions can influence outcomes and, consequently, recourse recommendations for others (many-to-one setting). 
Furthermore, individuals seeking recourse (i.e., \emph{seekers}) may engage with multiple providers (\Cref{fig:many-to-many}) to choose the most suitable among given recommendations.
The driving insight of this paper is that the effects of other stakeholders should not be ignored.

Recent studies have explored recourse frameworks that consider more than one seeker.
For instance, \citet{obrien_toward_2022} modified the recourse optimization problem, introducing \textit{Social-Welfare-Efficient Recourse} and \textit{Pareto-Efficient Recourse}.
Drawing upon game-theoretic principles, their illustrative use of the prisoner's dilemma highlights potential misaligned benefits in conventional recourse recommendations and provides a compelling ethical rationale for reconsidering algorithmic interventions in a many-to-one setting.
In a similar vein, \citet{fonseca_setting_2023} propose an agent-based simulation framework modeling multi-seeker-single-provider dynamic competition among recourse seekers over time, studying whether negatively classified instances will ever attain recourse in light of new individuals entering the seeker pool. 
Their work reveals that initially promising recourse recommendations may lose effectiveness due to continuous environmental changes.
This emphasizes the importance of accounting for both competition and temporal shifts when formulating interventions.
While these studies have extended the literature to settings with multiple recourse seekers, they continue to assume a single provider (\Cref{fig:many-to-one}). 

Also, while algorithmic recourse is typically framed as an individual intervention, providing a single affected person with minimal cost actions to reverse a fixed model decision, recourse objectives also implicitly induce a system-level allocation problem. 
In realistic, decision-making systems have limited capacity to provide favorable outcomes or actionable support, and multiple individuals may simultaneously seek recourse from multiple decision-makers. 
When such capacity constraints are present, recourse recommendations compete for scarce resources, transforming recourse from a purely individual optimization problem into a question of how limited intervention capacity should be allocated across a population. 
This observation aligns algorithmic recourse with broader questions of welfare maximization under resource constraints, a connection that has received comparatively limited attention in the recourse literature.

From this perspective, social welfare provides an objective for evaluating recourse systems at scale. 
Rather than considering recourse as an isolated right exercised independently by each individual, a welfare-based formulation evaluates how effectively a system converts limited intervention capacity into meaningful, low-cost opportunities for improvement across the population. 
Welfare-centric views have played a central role in the study of fairness in automated decision-making, where individual utilities, distributive justice, and aggregate outcomes are explicitly traded off \citep{binns_fairness_2017, hu_fair_2020, heidari_fairness_2018, corbett-davies_algorithmic_2017, bertsimas_efficiency-fairness_2012}. 
In our setting, we will characterize the best achievable population-level outcome under explicit capacity constraints. 
The resulting welfare gap quantifies the efficiency loss that arises when individually optimal recourse recommendations are infeasible at the system level.

Recent work has begun to demonstrate that recourse objectives themselves can induce inequitable or inefficient outcomes when optimized locally. 
In particular, \citet{perello_discrimination_2025} show that standard algorithmic recourse formulations may generate discrimination even in the absence of biased decision models, due to the interaction between recourse costs and population-level constraints. 
Beyond discrimination,~\citet{kugelgen_fairness_2022} show that cost-based recourse can embed structural inequities even when classifiers are causally fair, and~\citet{bell_fairness_2024} argue that minimizing individual recourse cost is insufficient when individuals face systematically different opportunity structures, further motivating our shift from individual cost minimization to population-level welfare optimization.
Related work in operational and algorithmic allocation studies how limited resources should be distributed fairly or efficiently across individuals with heterogeneous needs \citep{rea_unequal_2021,cousins_good_2024}. 
There remains a gap regarding algorithmic recourse in situations involving multiple recourse providers each potentially impacting outcomes with their own decision models.
Existing approaches typically overlook how providing recourse recommendations simultaneously to multiple recourse seekers can benefit society and overall recourse actionability through interactions among individuals. We model explicitly provider capacities and recourse costs jointly and expose a structural source of inefficiency that cannot be addressed by improving individual recourse methods alone.

Our work intersects with a broader literature on multi-agent resource allocation, where the
central challenge is distributing scarce resources among competing agents to maximize
collective welfare. 
\citet{dickerson_online_2019} study online bipartite
matching under capacity constraints in crowdsourcing and organ exchange markets,
demonstrating that system-level coordination consistently outperforms independent per-agent
assignment when resources are limited. 
In a related vein, \citet{li_incorporating_2019}
address weighted bipartite matching for kidney exchange, showing that incorporating all
compatible donor-recipient pairs through dynamic matching substantially improves social
welfare without sacrificing fairness for subpopulations. 
These works share our framing of capacity-constrained bipartite matching as a 
welfare optimization problem, but our framework is distinguished by the fact that the costs driving the matching are 
recourse costs, which introduces a direct coupling between the matching outcome and the validity of individual recourse recommendations, a connection we formalize 
in~\Cref{sec:many_to_many}.

The welfare gap we define also bears a conceptual resemblance to regret metrics in online
learning. 
In multi-armed bandit (MAB) problems, regret measures the cumulative gap between
an algorithm's realized reward and that of the best arm in hindsight~\citep{bubeck_regret_2012}.
Our welfare gap plays an analogous role: it quantifies the loss incurred when seekers act
in isolation under fixed capacities, relative to the optimal centrally coordinated outcome.
This connection is sharpest in the literature on fair multi-agent bandits. 
\citet{hossain_fair_2021} study the assignment of agents to arms to maximize Nash
social welfare, defining regret relative to the welfare-optimal policy. 
\citep{patil_achieving_2021} introduce fairness constraints requiring minimum pull
frequencies for each arm, directly analogous to our provider capacity constraints, and
show that fairness and efficiency can be simultaneously achieved with bounded regret.
Our welfare gap can be understood as a static, full-information counterpart to these
dynamic, partial-information regret notions. 
Both capture the same fundamental tension
between individual optimality and system-level feasibility, but in our setting the gap
arises from capacity misallocation rather than from learning under uncertainty.

\paragraph{Our Contributions}
We extend recourse to a many-to-many setting with capacity constraints, where the individually computed recourse recommendations no longer compose independently, and stakeholder interactions directly affect recourse validity.
In our framework we move past the unrealistic assumption of infinite provider capacity wherein seekers are able to match with any provider in absence of other simultaneous matches.
Our proposed formulation includes multiple recourse seekers and multiple recourse providers and examines how individual recommendations in such settings affect the overall system.
We formalize this interaction as a \emph{capacitated weighted bipartite matching problem} and determine optimal recourse outcomes using a linear-programming approach, thereby maximizing social welfare under capacity constraints.
Specifically, we evaluate the population cost or social welfare of recourse, summed over all individuals, under realistic capacity limits and contrast it with the case of unlimited provider capacity.

Further, we identify a welfare gap between the socially optimal solution, computed by a central planner, and the unrealistic individually optimal outcome, where each seeker acts in isolation and selects the provider offering the lowest recourse cost, without coordination or consideration of provider capacity constraints.
To minimize this gap, we introduce a second optimization layer that finds the best distribution for a total fixed capacity over providers.
We propose a systematic procedure to find this optimal distribution and minimize the gap.

Since algorithmic recourse methods provide recommendations that minimally change the initial situation to reach the favourable outcome, we add the third optimization layer that minimizes the welfare gap while penalizing deviations from the initial capacity values.
Solving this problem yields both the best provider capacities, accounting for the modification penalty, and the corresponding social welfare matching.

Finally, we extend our framework to include inequality-averse system objectives through a concave social-welfare formulation that prioritizes low-utility seekers under capacity constraints.

%% file: sections/Many-to-Many_Recourse_Optimization.tex
\section{Many-to-Many Recourse Optimization}
\label{sec:many_to_many}
In this section, we formalize the matching problem that serves as the foundation of our optimization framework.
The problem is modeled as a bipartite graph as shown in Figure~\ref{fig:many-to-many}, with loan seekers on one set and loan providers (e.g., banks) on the other, as follows:
\begin{itemize}
    \item{\textbf{Seekers} $\mathcal{S}$:} \( \{s_i ~|~ s_i \in \mathcal{S}, ~\forall~ i \in [n] \}\), each characterized by a feature vector \( x_i \).
    
    \item{\textbf{Providers} $\mathcal{P}$:} \( \{p_j ~|~ p_j \in \mathcal{P}, ~\forall~ j \in [m] \}\), each equipped with a classifier (w.l.o.g. binary model) $h_j$ to accept or reject the seekers and a matching capacity \( k_j \).
    All the seekers are initially rejected by all providers, i.e., 
    $
        h_j(x_i) = -1 ~ \forall ~ i, j,
    $
    meaning that each seeker will have a recommendation from all providers.
\end{itemize}
\noindent Furthermore, it is assumed that a \textit{central planner} will coordinate matches between seekers and providers (i.e., \cref{equ:first-optimization}) and potentially redistribute existing capacity among providers (i.e., \cref{equ:second_optimization} and \cref{equ:third_optimization}).
\paragraph{Recourse cost computation}
%
The standard algorithmic recourse setting assumes that a seeker (e.g., \(s_i\)) seeks to obtain recourse recommendations from a provider (e.g., \( p_j \)).
%
This minimal change, as defined by \citet{ustun_actionable_2019}, is defined as the minimal effort or cost required to change an individual's input features such that a predictive model will reverse its output from an undesirable outcome to a desirable one. 
Formally, given provider \(p_j\)'s decision model \(h_j\) and an input feature vector \(x_i\) corresponding to the characteristics of seeker \(s_i\), such that \(h_j(x_i) = -1\) (assumed binary w.l.o.g.), the recourse cost is defined as the solution to the following optimization problem:
\[
    c_{ij} =  \min_{a \in A(x_i)} \text{cost}(a; x_i) \quad \text{s.t.} \quad h_j(x_i + a) = +1\quad \forall ~ i,j
\]
\noindent where \( a \) is an action vector representing feasible changes to the features of \( x_i \), and \( A(x_i) \) is the set of allowed actions based on domain constraints (e.g., mutability and bounds on feature changes). The function \( \text{cost}(a; x_i) \) quantifies the difficulty of applying action \( a \) to instance \( x_i \).  
If a feasible action exists that satisfies the constraint, the minimal value of \( \text{cost}(a; x_i) \) is the \emph{recourse cost}. Therefore, \(c_{ij} \) represents the minimal change required for seeker \(s_i\) to achieve approval from provider \(p_j\).
Our proposed framework is agnostic to the choices of recourse method and providers' model, operating only on the minimum cost of change required for each pair of seeker \(s_i\) and provider \(p_j\).
\paragraph{Bipartite Graph Construction.}
Once all minimal recourse costs, $c_{ij}$, between seekers and providers are precomputed, we construct a \emph{weighted bipartite graph} \(\mathcal{G} = (\mathcal{V}, \mathcal{W})\), where nodes \(\mathcal{V} = \mathcal{S} \cup \mathcal{P}\) and \[ \label{equ:exp_weight_transform}
    \mathcal{W} := \left\{ w_{ij} \;\middle|\; w_{ij} = e^{-\gamma \cdot c_{ij}}, ~ \forall ~ i, j \right\}.
\]
%

where $\gamma > 0$ is a scaling parameter controlling the sensitivity of the transformation. 
This exponential transformation converts costs into edge weights, enabling algorithms such as the maximum-weight bipartite matching~\citep{kuhn_hungarian_1955} to prioritize low-cost (i.e., 
efficient) recourse assignments while maximizing overall match coverage.
Furthermore, this transformation is monotone in $c_{ij}$ and consistent with an exponential utility representation exhibiting Constant Absolute Risk Aversion (CARA), a standard model in economics for capturing diminishing marginal disutility of cost or effort~\citep{arrow_theory_1971, pratt_risk_1992}: lower recourse costs yield higher utility, while differences among low-cost interventions are weighted more heavily than differences among high-cost ones. 
Beyond its utility interpretation, $\gamma$ represents a meaningful modeling decision about how strongly cost differences between providers are reflected in the matching weights. 
%
%
A formal analysis of $\gamma$'s effect on social welfare is provided in~\Cref{sec:sensitivity}.

\paragraph{Optimization Model}
Next, with consideration for real-world constraints on provider capacity, we formulate a \emph{capacitated weighted bipartite matching problem}. 
If we denote the maximum weight for seeker \( i \) accordingly as \(w^*_i~=~\max_{j} (w_{ij})\)  (corresponding to the least costly recommendation received), we can then measure the ideal scenario in which each seeker attains its optimal outcome, independently of other seekers, as:
\begin{equation}
    \text{\color{gray} Individual Welfare} := \sum_{i=1}^{n} w^*_i
\end{equation}    
%
However, this ideal scenario assumes that providers have unbounded capacity (w.l.o.g., at least the number of seekers for each provider), meaning that they can freely provide the resources, which is not realistic.
In practice, each provider has a limited capacity \( k_j \), meaning they can serve only a finite number of seekers.
Taking a systems-level view and aiming to minimize the overall cost of recourse across all seekers and providers,\footnote{This assumes that providers do not have ulterior preferences affecting the matching process.} the optimal matching under capacity constraints is determined as:
%
\begin{equation}
    \text{\color{gray} Social Welfare} := \max_{z_{ij}} \sum_{i=1}^{n} \sum_{j=1}^{m} w_{ij}\, z_{ij}.
\end{equation}
where \( z_{ij} \) are binary decision variables that indicate whether seeker \( i \) is assigned to provider \( j \).
This equation has the structure of a classical capacitated bipartite matching problem, solvable in polynomial time via assignment algorithms~\citep{kuhn_hungarian_1955}.
In our implementation, to obtain the optimal matching, we encode the above formulation as a mixed-integer linear program (MILP) and solve it with the Gurobi Optimizer~\citep{gurobi_optimization_llc_gurobi_2024}\footnote{ Gurobi’s branch-and-bound engine—augmented with presolve, cutting-plane generation, and heuristic warm-starts--guarantees global optimality.}~as follows:
\begin{equation}
\begin{aligned}
        \text{\color{gray} Social Welfare} := \max_{z_{ij}} \quad &\sum_{i=1}^{n} \sum_{j=1}^{m} w_{ij}\, z_{ij} \\
        \text{s.t.} \quad &\sum_{j=1}^{m} z_{ij} \leq 1 \quad \forall i, \quad \sum_{i=1}^{n} z_{ij} \leq k_j \quad \forall j, \quad z_{ij} \in \{0, 1\} \quad \forall i,j\\
        &\underbrace{\hphantom{\sum_{j=1}^{m} z_{ij} \leq 1 \quad \forall i}}_{\text{\color{gray}Matching Constraint}} \quad \underbrace{\hphantom{\sum_{i=1}^{n} z_{ij} \leq k_j \quad \forall j}}_{\text{\color{gray}Capacity Constraint}} \quad \underbrace{\hphantom{z_{ij} \in \{0, 1\} \quad \forall i,j}}_{\text{\color{gray}Edge Constraint}}
\end{aligned}
\label{equ:first-optimization}
\end{equation}
Under unbounded provider capacity, the optimal matching over $z_{ij}$ is achieved when each seeker is matched with the provider that minimizes its individual recourse cost.
However, capacity constraints may leave some seekers to accept suboptimal assignments ($w_{ij} < w^*_i$), resulting in higher recourse costs than their ideal match.
This discrepancy is quantified by the gap between the ideal individual welfare and the realized social welfare:
\begin{equation}
    \text{\color{gray} Welfare Gap} := \left( \sum_{i=1}^{n} w^*_i \right) - \left( \sum_{i=1}^{n} \sum_{j=1}^{m} w_{ij}\, z_{ij} \right).
\end{equation}
This gap highlights a critical design challenge: given a fixed total amount of provider capacity, how should these limited resources be distributed across providers to minimize the welfare gap?
A naive uniform distribution of provider capacities may lead to significant welfare losses. 
In contrast, allocating more capacity to providers most preferred by seekers, i.e., those associated with lower recourse costs, can substantially reduce the welfare gap, even when the total capacity remains fixed.
In the following section, we introduce a systematic approach that not only identifies distinct allocation scenarios based on varying resource availability but also provides an optimization approach to allocate capacity effectively and minimize the welfare gap.

%% file: sections/Minimize_Welfare_Gap.tex
\section{Minimizing the Welfare Gap}
\label{sec:minimize-gap}
\begin{figure}[t]
    \vskip 0.2in
    \begin{center}
    \Description{Welfare-gap plot}
    \centerline{\includegraphics[width=0.5\columnwidth]{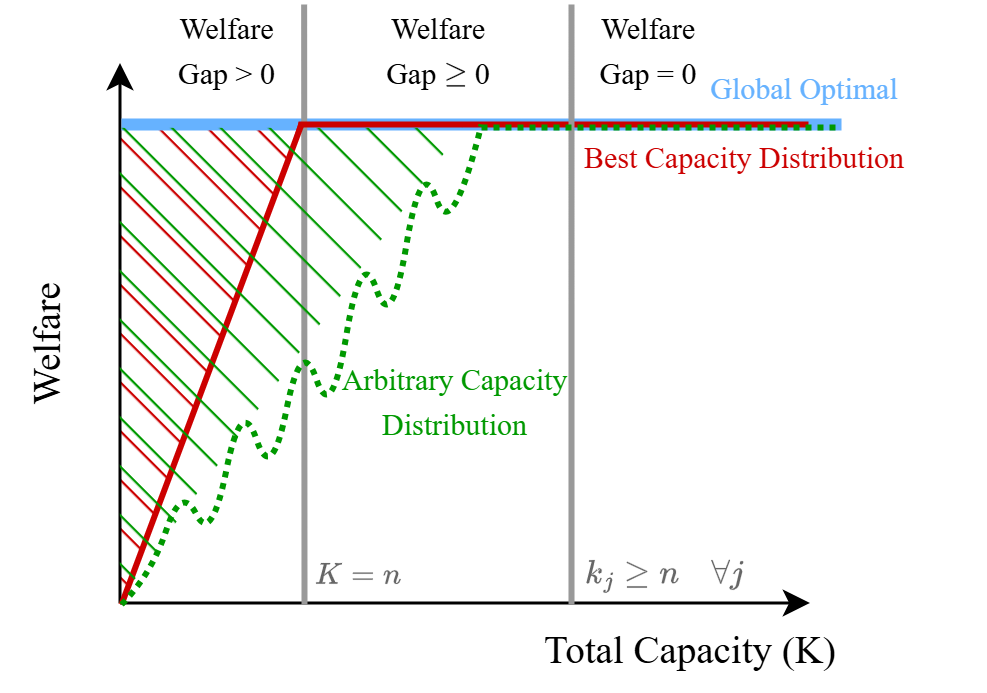}}
    \caption{The effect of capacity distribution on social welfare outcomes. The pink dashed plot (with shaded region) shows the welfare achieved under an arbitrary capacity allocation (\Cref{equ:first-optimization}), which may result in a non-zero welfare gap even when sufficient total capacity is available. The dark red plot represents the optimized distribution obtained by solving~\Cref{equ:second_optimization}, which maximizes social welfare under a total capacity constraint. The blue horizontal line shows the global individual-optimal welfare, attainable only when total capacity is unconstrained. The gap between the pink and red curves highlights the inefficiency introduced by uncoordinated capacity allocations.}
    \label{fig:welfare-gap}
    \end{center}
    \vskip -0.2in
\end{figure}

Under a fixed total provider capacity $K = \sum_{j=1}^{m} k_j$, the welfare gap can vary depending on how capacity is distributed among providers. 
In fact, for any given $K$, there is an optimal allocation of provider capacities \(k_j\) that minimizes this welfare gap. 
This observation leads to a new optimization problem involving two sets of decision variables namely, the integer variables \(k_j~\forall~j\), representing provider capacities in the optimal solution, and the matching variables \(z_{ij}\), as previously defined, indicating the best matching under the system settings. 
\begin{equation}
    \begin{aligned}
        \max_{z_{ij},~ k_j}            \quad &\sum_{i=1}^{n} \sum_{j=1}^{m} w_{ij}\, z_{ij} \quad
        \text{s.t.}
                        &\sum_{j=1}^{m} k_j = K \quad \forall \, j \quad \text{\color{gray} Total Capacity Constraint}
    \end{aligned}
    \label{equ:second_optimization}
\end{equation}
The matching, capacity, and edge constraints remain the same as before in~\Cref{equ:first-optimization}, with the additional constraint on the total capacity.
\Cref{fig:welfare-gap} represents how this welfare gap varies according to the total available resources and their distribution across providers.
In particular, the figure highlights two notable cutoffs and three distinct areas.
In the first area, where the total available resources are fewer than the number of seekers, i.e., \(K = \sum_{j=1}^{m} k_j < |S|\), a welfare gap greater than zero is inevitable.
Under these circumstances, some seekers will remain unmatched, and the depicted minimum gap represents the best achievable outcome through optimal capacity distribution.
The first critical point is reached once available resources equal the number of seekers, where the welfare gap \emph{can} reach zero \emph{if} the distribution of these resources could be optimized and aligned with the best choice of each seeker. 
At the second critical point, where each provider individually possesses resources equal to the number of seekers, the welfare gap is guaranteed to be zero as seekers can freely match with their preferred providers.
In the area between the two critical points, while resources are plentiful enough for a zero welfare gap distribution over capacities to exist, any suboptimal distribution of capacities among providers will result in a welfare gap greater than zero.
Therefore, identifying the optimal capacity distribution becomes an essential challenge, specifically, determining how capacities can best be allocated to minimize the welfare gap given fixed total resources.

To determine the optimal distribution of provider capacities,~\Cref{equ:second_optimization} could again be solved using a Mixed-Integer Linear Programming (MILP) method with solvers such as Gurobi~\citep{gurobi_optimization_llc_gurobi_2024}. 
However, the structure of the matching weights reveals two key patterns, suggesting an approach simpler than solving an MILP directly. 
In~\Cref{alg:optimal_capacity}, we propose a systematic method that assigns capacities based on each seeker's top-ranked matching weight, previously defined in~\Cref{sec:many_to_many} as \(w^*_i\). 
For each seeker \(i\), the value \(w^*_i\) represents the maximum possible contribution of that seeker to the total welfare, as no assignment can surpass this highest-weight edge.
Moreover, since at most \(K\) seekers can receive recourse, excluding any of the top \(K\) highest-weight edges directly reduces the achievable welfare. 
Thus, the welfare of any feasible solution is bounded above by \(\sum_{i \in \mathcal{S}_K} w_i^*\), where \(\mathcal{S}_K\) denotes the set of seekers corresponding to the \(K\) highest-ranked edges, and \(j_i^* = \arg\max(w_i^*)\) denotes the index of the provider most preferred by seeker \(s_i\). Then the capacity vector \(k^*\) is defined by counting how often each provider appears among the top-K individually preferred matches (e.g. \(w^*_i\)). 
\begin{align*}
k^*_j = \left|\{i \in S_K \mid j_i^* = j \} \right|, \quad j = 1, ..., m
\end{align*}
These insights lead to~\Cref{alg:optimal_capacity}, which offers a more direct and efficient alternative to MILP. 
After computing all \(w^*_i\) values in \(O(nm)\) time (\(n\) seekers and \(m\) providers), we sort the resulting values once in \(O(n\log(n))\) time.
Capacities are then assigned to providers precisely according to the top \(K\) dominant edges.
This approach ensures provider capacities align directly with seeker preferences, thereby naturally minimizing the welfare gap by emphasizing matches that are both individually optimal and beneficial for the overall system.
Although, ~\Cref{equ:second_optimization} could be solved using a solver such as Gurobi,~\Cref{alg:optimal_capacity} exploits the special structure of the problem to provide a direct, closed-form solution and it recovers the same optimal social welfare as the solution from the solver.
\begin{algorithm}[t]
  \caption{Optimal Capacity Distribution}
  \label{alg:optimal_capacity}
  \begin{algorithmic}[1]
    \STATE \textbf{Input:} seekers $\mathcal{S}$, providers $\mathcal{P}$, weights $w_{ij}$, total capacity $K$
    \STATE \textbf{Output:} provider capacities $k = (k_1,\dots,k_{|\mathcal{P}|})$
    \STATE Initialize empty list $\mathcal{L}$
    \FOR{each seeker $i \in \mathcal{S}$}
        \STATE $w_i^* = max_j(w_{ij})$ \hfill
        \STATE $j_i^* \leftarrow \arg\max_{j \in \mathcal{P}} \; w_i^*$ \hfill\COMMENT{best provider for seeker $i$}
        \STATE Append triple $(i,\; j_i^*,\; w_i^*)$ to $\mathcal{L}$
    \ENDFOR
    \STATE Sort $\mathcal{L}$ in descending order of weight
    \STATE Select the first $K$ triples of $\mathcal{L}$ \hfill\COMMENT{top‐$K$ matches}
    \STATE Initialize $k_j \leftarrow 0$ for all $j \in \mathcal{P}$
    \FOR{each selected triple $(i,\; j,\; w)$}
        \STATE $k_j \leftarrow k_j + 1$
    \ENDFOR
    \STATE \textbf{return} capacity vector $k$
  \end{algorithmic}
\end{algorithm}
\subsection{Capacity Redistribution with Penalizing Modifications}
\label{sec:3.1}
While the previous section addressed how provider capacities can be optimally distributed to minimize the welfare gap for a given total system capacity \(K\), additional considerations may be warranted. 
In practice, recourse methods often operate within established configurations determined by existing organizational structures, resource availability, and operational constraints.
Transitioning from the current provider capacity configuration to an optimal setup typically involves real-world adjustment costs.
Simply recommending an entirely new capacity distribution may be impractical or expensive to implement.
Therefore, even though optimally reallocating capacities reduces the welfare gap, the costs of these adjustments must be carefully balanced against minimizing the welfare gap.
To address this challenge, we extend our optimization framework by explicitly penalizing deviations from the initial capacities.

Let \(\hat{k_j}\) denote the initial capacity of provider \(j\), and \(\tilde{k_j}\) represent the target capacity after configuration. 
The change in capacity \(\Delta k_j = \tilde{k_j} - \hat{k_j}\) can penalize large changes with \(\beta_j  |\Delta k_j|\), where \(\beta_j\geq0\) controls the penalty sensitivity for each of the providers accordingly. 
%
%
Integrating this penalty into our optimization leads to a multi-objective problem, balancing social welfare maximization with minimization of capacity adjustment penalties. The modified objective function is:
\begin{equation}
\begin{aligned}
    \text{Welfare} = \max_{z_{ij},~ k_j} \left( \sum_{i=1}^{n} \sum_{j=1}^{m} w_{ij} z_{ij} - \sum_{j=1}^{m} \beta_j|\Delta k_j| \right)
\end{aligned}
\label{equ:third_optimization}
\end{equation}

\noindent subject to the same matching, capacity, total capacity, and edge constraints previously defined in~\Cref{equ:second_optimization}. 
Unlike~\Cref{equ:first-optimization}, the penalized formulation no longer reduces to a standard bipartite matching or flow problem, due to the interaction between capacity constraints. 
As a result, this enhanced formulation constitutes a MILP problem, solvable by recent versions of solvers like Gurobi~\citep{gurobi_optimization_llc_gurobi_2024}.
Ultimately, this optimization simultaneously identifies the optimal matching \(z_{ij}\) and optimal provider capacities \(k_j\), clearly indicating how capacities should be adjusted from their initial configuration. 
The resulting solution provides a practical recourse recommendation system effectively balancing improvements in social welfare with realistic operational constraints.

%% file: sections/Fairness.tex
\section{ Inequality-Averse Recourse Optimization}
\label{sec:fairness}
We have introduced the three-layer optimization framework in \Cref{sec:many_to_many} under a utilitarian objective, which is maximizing the aggregate sum of weights across the population. 
While the approach in \Cref{sec:minimize-gap} minimizes the total welfare gap, it is theoretically blind how that welfare is distributed among individuals. 
In resource-constrained settings, a purely utilitarian objective naturally favors \emph{efficient} matches, seekers with low recourse costs (high edge weights \(w_{ij}\)), while it may inadvertently disadvantage those with higher costs. 
This creates a risk that efforts to maximize overall performance systematically disadvantage the most vulnerable seekers.
If a recourse system achieves high total welfare by serving only the easiest cases while leaving others with no actionable recommendation, it fails to provide the meaningful opportunities required by the right to recourse. 
To address this, we extend our framework to explicitly model \emph{distribution-sensitive welfare}, trading off allocative efficiency for distributional equity~\citep{bertsimas_efficiency-fairness_2012, hu_fair_2020}.

To account for inequality aversion in the matching process, we use a concave social welfare function. 
Unlike the utilitarian approach in \Cref{equ:first-optimization}, which assigns constant marginal value to utility gains, a concave objective places diminishing marginal value on increasing the utility o well-off seekers. 
This functional form prioritizes improvements for those with the lowest weights, a standard approach in welfare economics for modeling inequality aversion~\citep{atkinson_measurement_1970, arrow_theory_1971}.

Let $u_i = \sum_{j=1}^m w_{ij} z_{ij}$ denote the utility attained by seeker $i$ under a given matching. 
We replace the linear objective with the following concave welfare function, consistent with Constant Absolute Risk Aversion (CARA) assumptions~\citep{pratt_risk_1992}:
\begin{equation}
    \max_{z_{ij}} \sum_{i=1}^n (u_i)^{\alpha}, \quad \text{where } \alpha \in (0, 1)
    \label{eq:concave_obj}
\end{equation}
subject to the matching and capacity constraints defined in \Cref{equ:first-optimization}. 
The parameter $\alpha$ serves as a coefficient of \emph{inequality aversion}. 
When $\alpha=1$, the objective recovers the standard utilitarian function. 
As $\alpha \to 0$, the optimization approaches the Rawlsian Max-Min principle~\citep{rawls_theory_2003}, prioritizing the maximization of the worst-off seeker's utility above all else. 
This formulation provides a principled mechanism to balance aggregate efficiency with equity without requiring rigid constraints or manual quotas.

To evaluate the fairness of allocations produced by our framework, we define a specific Equity metric, denoted as $\mathcal{E}$. 
While the social welfare shows the system's total efficiency, $\mathcal{E}$ quantifies the Rawlsian difference principle by measuring the \emph{floor} of the recourse distribution. 
We define $\mathcal{E}$ as the minimum utility attained by any successfully matched seeker: 
\begin{equation} 
    \mathcal{E} = \min_{i \in \mathcal{M}} u_i = \min_{i \in \mathcal{M}} \left( \sum_{j=1}^m w_{ij} z_{ij} \right) 
\label{eq:equity_metric}
\end{equation} 
where $\mathcal{M} = \{i \mid \sum_j z_{ij} = 1\}$ represents the set of matched seekers. 
An increase in $\mathcal{E}$ indicates that the system has successfully increased the outcome for the least advantaged individual. 
In~\Cref{sec:equity_experiment}, we use this metric to demonstrate that our concave formulation can significantly increase $\mathcal{E}$ with only marginal costs to social welfare.

%% file: sections/Experiments.tex
    \section{Experiments}
    \label{sec:experiments}
This section demonstrates how the proposed many-to-many recourse framework
behaves in practice.
All experiments are fully reproducible from the scripts accompanying this paper.

\paragraph{Datasets and Models}
We evaluate our framework on three datasets across two experimental scales.
For small-scale mechanistic illustration we use the synthetic \textbf{Two-Moon}
benchmark, whose non-linear geometry yields heterogeneous classifiers with
disagreements on decision boundaries.
For large-scale validation we use the real-world \textbf{Credit}~\citep{i-cheng_yeh_default_2009}
and \textbf{COMPAS}~\citep{larson_propublica_2016} datasets, widely used in recourse research.
 
In the \emph{small-scale} setting, four classification models serve as the $m=4$ providers: logistic regression, a multilayer perceptron (MLP), a decision tree, and a random forest.
We subsample $n=8$ seekers for Two-Moon, instances that are rejected by all providers,
forming the pool $\{s_i\}_{i=1}^{8}$ used for visualization in~\Cref{fig:exp-sixpanel}.
 
In the \emph{large-scale} setting, we construct $m=15$ providers spanning different performance across the same four model families.
Provider heterogeneity is introduced by varying model-specific hyperparameters
(e.g., regularization strength, tree depth, number of estimators), yielding
providers with genuinely different decision boundaries and predictive behaviors (\Cref{tab:provider_performance} shows the preformance of these providers' model in~\Cref{app:provider_perf}).
Additionally, as the number of providers grows, finding seekers rejected by all providers becomes difficult, since any single permissive provider can disqualify a candidate from the seeker pool. 
To ensure a sufficient pool of seekers rejected by all providers, we apply provider-specific classification thresholds: some providers use more conservative (higher) decision thresholds, increasing the instances they reject and thereby enlarging the pool of seekers who are denied by all $m=15$ providers.

\paragraph{Counterfactual Generation}
Our framework is agnostic to the choice of recourse method, operating only on
the minimum cost $c_{ij}$ for each seeker–provider pair, precomputed using a recourse method of choice.
We exploit this agnosticism by using two recourse methods across our settings.
For the small-scale Two-Moon experiments we use Model-Agnostic Counterfactual Explanations (MACE)~\citep{karimi_model-agnostic_2020}, and for the large-scale Credit and COMPAS experiments we use Diverse Counterfactual Explanations (DiCE)~\citep{mothilal_explaining_2020}.
We use the distance norm $\ell_1$ for DiCE experiments and $\ell_\infty$ for MACE as the recourse cost measure.
For every pair $(s_i, p_j)$ the recourse method outputs a counterfactual
instance together with its cost $c_{ij}$.
This yields $8\times4=32$ weights for Two-Moon and $200\times15=3{,}000$ weights for each of Credit and COMPAS.

\paragraph{Optimization Setup}
The optimization proceeds in three layers across all experiments using an off-the-shelf MILP solver~\citep{gurobi_optimization_llc_gurobi_2024}.
\emph{(i)} Given a total capacity $K$ equal to the number of seekers and an
initial capacity vector $\hat{k}$,\footnote{For the small-scale setting the initial capacity vector is drawn uniformly at random; for the large-scale setting it is drawn from a Poisson distribution, reflecting a more realistic scenario of unequal baseline capacities whereby most providers have small capacity and some providers have large capacity.} \Cref{equ:first-optimization} finds the matching with the maximum social welfare under specified constraints.
\emph{(ii)}~\Cref{alg:optimal_capacity} solves~\Cref{equ:second_optimization}, tracing the full welfare curve as
$K$ ranges from $0$ to $n\cdot m$ and validating the theoretical cutoffs of~\Cref{sec:minimize-gap}.
\emph{(iii)} solves~\Cref{equ:third_optimization},
jointly optimizing matching and capacities while penalizing deviations from the initial vector.
Hyperparameter selection is discussed in Section~\ref{sec:sensitivity}; the
values used here are $\gamma=10$, $\beta=0.03$ for Two-Moon (MACE) and
$\gamma=100$, $\beta=0.15$ for Credit and COMPAS (DiCE).
\paragraph{Metrics} To evaluate the proposed framework, we introduce the \emph{Welfare Attainment Ratio}, 
defined as:
\begin{equation}
    \frac{\text{SW}}{\text{IW}} = \frac{\max_{z_{ij}} 
    \sum_{i=1}^{n} \sum_{j=1}^{m} w_{ij}\, z_{ij}}{\sum_{i=1}^{n} 
    \max_{j} w_{ij}}
    \label{eq:sw_iw}
\end{equation}
where the Social Welfare is feasible under constraints defined in~\Cref{equ:first-optimization}.
This ratio measures the fraction of the theoretically optimal individual welfare that the system achieves under capacity constraints. 
A value of 1 indicates that the system delivers the same aggregate welfare as if every seeker were matched to their highest-weight provider with no capacity limitations. 
A value below 1 quantifies the welfare loss introduced by 
capacity constraints and their distribution across providers. 
%

\paragraph{Results}
\input{sections/two-moons-results}
\Cref{fig:exp-C} illustrates the social welfare achieved with the capacity distribution returned by \Cref{alg:optimal_capacity} for each total capacity \(K\) as it increases.
The dashed red curve rises monotonically and meets the individual‐welfare upper bound (solid blue) exactly at the point \(K = n\).  
Beyond that, further capacity is non-essential, confirming the welfare gap would be zero in the case of finding the optimized distribution of capacities.

\Cref{fig:exp-D} for Two-Moon dataset fixes \(K = n\) (with \(n = 8\)) and starts from an arbitrary reference capacity vector \( \hat{\mathbf k}=(2,4,1,1)\); the first stage of our framework then computes the best matching for this initial configuration as shown in the figure.
Although every seeker is matched, the allocation falls short of the global individual welfare.
\Cref{fig:exp-B} exposes the source of this gap.
The next step of the algorithm finds each seeker’s best edge \(w_i^*\); distributing capacity according to \Cref{alg:optimal_capacity} and finds
\( \mathbf k^*=(0,2,4,4)\).  
With this distribution, the system reaches the welfare upper bound, which closes the gap entirely.
\Cref{fig:exp-F} shows the outcome of the cost-aware optimization
that penalizes deviations from \(\hat{\mathbf k}\).
The solver selects a softer move,
\( \mathbf{k} =(1,3,1,3)\) yet the matching in this setting retains 0.9938 of the global optimum.  
Hence, almost the full efficiency gain is
achievable with limited changes, demonstrating the practical value of our proposed framework.(\Cref{app:extra-exp} shows result of the same setting experiment with $\ell_1$ distance norm.)

To extend our evaluation, we summarize results for all datasets in \Cref{tab:l_infty_results}, which highlights the welfare attainable ratio for each optimization setting.
For the Credit and COMPAS datasets, the best matching from \Cref{equ:first-optimization} attained a social welfare that constitutes 0.841 and 0.832 of the individual welfare, respectively. 
The cost-aware optimization in \Cref{equ:third_optimization} selects a softer redistribution that preserved 0.948 and 0.931 of the optimal welfare for Credit and COMPAS datasets while making a more balanced capacity modification, compared to the sharper concentration found in \Cref{equ:second_optimization}. 
To the best of our knowledge, this is the first work that systematically aims to close the social welfare gap by moving beyond individual recourse recommendations and instead considering coordinated, system-level optimization across multiple seekers and providers. 
These results align with the findings from the synthetic dataset, reinforcing the effectiveness of targeted redistribution and demonstrating that even for complex, real-world data, high social welfare can be achieved with moderate system changes.
\begin{table}[t]
\centering
\caption{Comparative Welfare attainable ratio (SW/IW) on three different datasets of the three optimization formulations.}
\label{tab:l_infty_results}

\small
\setlength{\tabcolsep}{6pt}
\renewcommand{\arraystretch}{1.1}

\begin{tabular}{lccc}
\toprule
 & \textbf{Two-Moon} & \textbf{Credit} & \textbf{COMPAS} \\
\midrule
\Cref{equ:first-optimization} 
 & 0.931 
 & 0.841 
 & 0.832 \\

\Cref{equ:second_optimization} 
 & 1.000 
 & 1.000
 & 1.000 \\

\Cref{equ:third_optimization} 
 & 0.993 
 & 0.948 
 & 0.931 \\

\bottomrule
\end{tabular}

\end{table}

\subsection{Hyperparameter Sensitivity Analysis}
\label{sec:sensitivity}

\paragraph{Sensitivity to $\alpha$}
\label{sec:equity_experiment}
We evaluate the impact of the concave social welfare formulation (\Cref{sec:fairness}), focusing specifically on the COMPAS dataset which offers a more complex feature space with significant cost disparities between providers, in contrast to the Two-moon and Credit datasets. 
This structural heterogeneity is essential for this experiment, as it creates the necessary conditions to observe a meaningful trade-off between aggregate efficiency and individual equity.

To test the framework under these conditions, we use a high sensitivity parameter ($\gamma=50$), which further amplifies the utility gap between preferred and secondary providers. 
We vary the inequality aversion parameter $\alpha$ from $0.01$ (highly equity-focused) to $1.0$ (utilitarian) and measure system performance using two competing metrics, Social Welfare (aggregate efficiency) and the Equity Metric $\mathcal{E}$ (\Cref{eq:equity_metric}).

\begin{figure*}[t]
    \centering
    \Description{Experiments on efficiency and equity}
    \begin{subfigure}[b]{0.48\textwidth}      
        \centering
        \includegraphics[height=5cm]{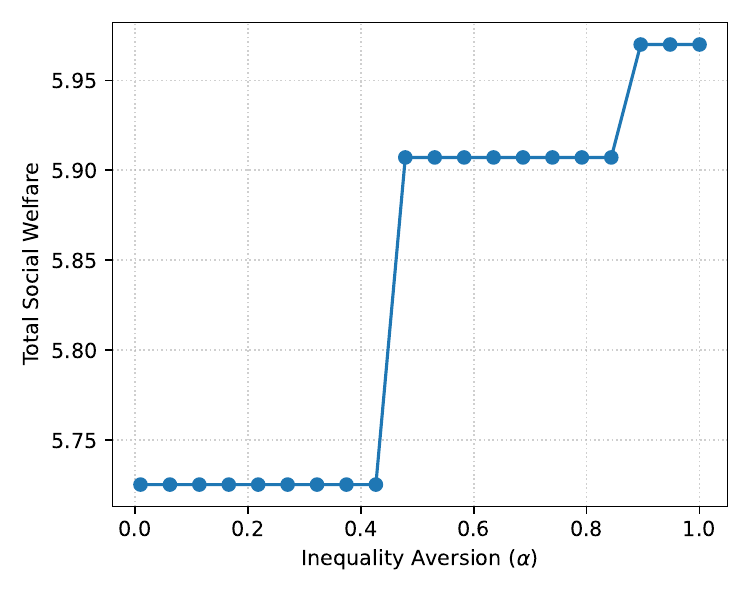}
        \caption{Efficiency Cost}
        \label{fig:exp-eff}
    \end{subfigure}
    \hfill
    \begin{subfigure}[b]{0.48\textwidth}   
        \centering
        \includegraphics[height=5cm]{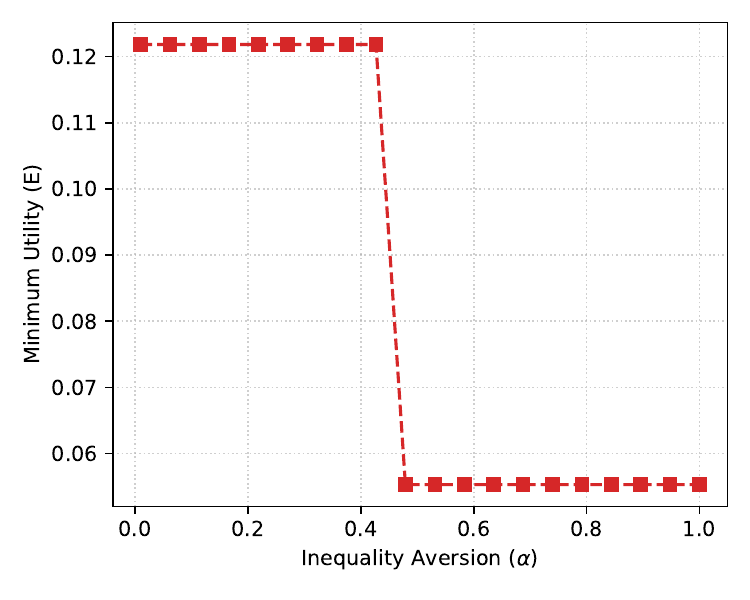}
        \caption{Equity Gain}
        \label{fig:exp-equity}
    \end{subfigure}

    \caption{Trade-off analysis on the COMPAS dataset with high sensitivity ($\gamma=50$). (a) Total Social Welfare shows a minor decline as the system becomes more fair (lower $\alpha$). (b) The Equity Metric $\mathcal{E}$ (Minimum Utility) significantly improves, validating that the concave formulation effectively protects the worst-off seekers.}
    \label{fig:equity_tradeoff}
\end{figure*}

The results, illustrated in~\Cref{fig:equity_tradeoff}, demonstrate a trade-off between these objectives. 
Under the standard utilitarian baseline ($\alpha=1.0$), the system achieves its maximum aggregate efficiency ($\approx 5.95$) but leaves the most disadvantaged seekers with a low utility baseline ($\mathcal{E} \approx 0.06$). 
As we introduce inequality aversion by decreasing $\alpha$ toward $0.01$, the system reallocates capacity to protect these vulnerable individuals.

This redistribution results in a monotonic but minor reduction in social welfare, which drops to approximately $5.75$, a relative efficiency loss of only 3.4\%. 
In contrast to this marginal cost, the gain in equity is substantial, the minimum matched utility $\mathcal{E}$ doubles from $\approx 0.06$ to $\approx 0.12$. 
This finding validates that the concave welfare formulation successfully elevates the recourse floor for the worst-off seekers without causing a significant burden on overall system efficiency.

\begin{figure}[t]
    \centering
    \begin{subfigure}[b]{0.48\columnwidth}
        \centering
        \includegraphics[width=\linewidth]{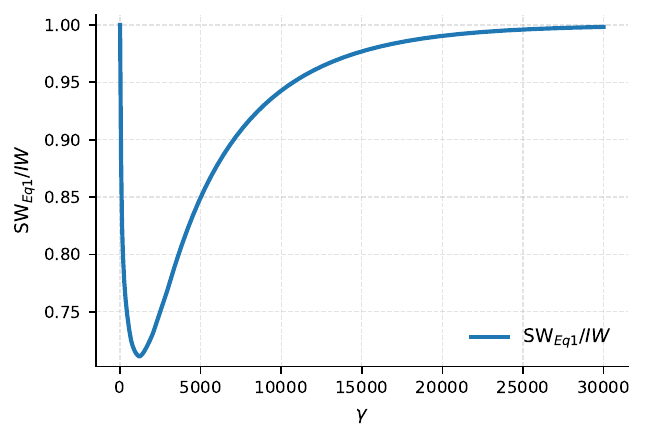}
        \caption{Sensitivity to $\gamma$}
        \label{fig:gamma_sensitivity}
    \end{subfigure}
    \hfill
    \begin{subfigure}[b]{0.48\columnwidth}
        \centering
        \includegraphics[width=\linewidth]{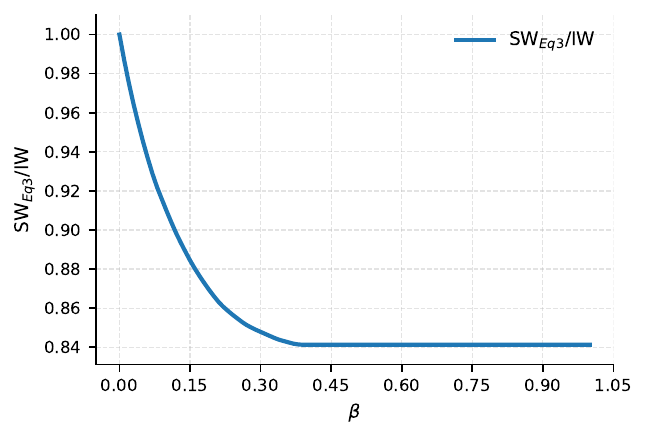}
        \caption{Sensitivity to $\beta$}
        \label{fig:beta_sensitivity}
    \end{subfigure}
    \caption{Hyperparameter sensitivity analysis on the large-scale 
Credit setting ($n=200$, $m=15$). 
(a) As $\gamma$ increases, 
$\text{SW}_{\text{Eq.1}}/\text{IW}$ exhibits a U-shaped trend, 
initially decreasing due to capacity-induced displacement and 
later recovering as the matching becomes dominated by maximum-weight 
edges. The stable elbow at $\gamma \in [50, 150]$ motivates our choice 
of $\gamma=100$. 
(b) $\text{SW}_{\text{Eq.3}}/\text{IW}$ 
interpolates smoothly between Equation~\eqref{equ:second_optimization} 
at $\beta=0$ and Equation~\eqref{equ:first-optimization} at $\beta \approx 0.40$.}
    \label{fig:sensitivity}
\end{figure}

\paragraph{Sensitivity to $\beta$.}
The penalty coefficient $\beta$ in~\Cref{equ:third_optimization} controls the trade-off between welfare maximization and deviation from the initial 
capacity configuration. \Cref{fig:sensitivity} shows how SW/IW 
evolves as $\beta$ increases from 0 to 1 on the large-scale experiment setting. 
At $\beta = 0$, the penalty term vanishes and~\Cref{equ:third_optimization} reduces exactly to~\Cref{equ:second_optimization}), recovering the individual welfare completely.
As $\beta$ increases, the growing penalty discourages capacity modifications, and $SW/IW$ decays monotonically until 
converging to~\Cref{equ:first-optimization}), where no capacity redistribution takes place. 
This confirms that $\beta$ provides a smooth 
interpolation between the two extremes of the framework, and that practitioners can select any value in $[0, 1]$ to trade off welfare gain against the cost of capacity modification.

\paragraph{Sensitivity to $\gamma$.}
The scaling parameter $\gamma$ controls how strongly cost differences 
between providers are reflected in the matching weights $w_{ij} = 
e^{-\gamma c_{ij}}$.~\Cref{fig:sensitivity} shows how the attainable welfare ratio $SW/IW$
varies as $\gamma$ ranges from 0 to 15,000 in the large-scale 
experiment settings, revealing a characteristic U-shaped curve with 
three distinct regimes.

At $\gamma = 0$, all weights equal 1 and the matching is purely 
capacity-driven, treating every seeker-provider pair as equally 
desirable regardless of cost; $SW/IW$ reflects capacity utilization 
alone. Since $K = n$, every seeker is matched and $SW/IW = 1$.

As $\gamma$ increases from 0, weights become heterogeneous, and the 
matching starts favoring high-weight assignments. However, capacity 
constraints prevent some seekers from accessing their preferred 
provider. This gap grows exponentially with $\gamma$, 
pulling $SW/IW$ downward. The minimum value observed around 
$\gamma \approx 1{,}000$ in~\Cref{fig:sensitivity}, represents 
the worst-case welfare loss under the initial capacity vector $\hat{k}$: 
the point where cost discrimination is strong enough to create 
significant displacement losses but the dominant term has not yet taken 
over. The depth of this minimum characterizes how poorly $\hat{k}$ is 
aligned with the cost structure of the system.

At very large $\gamma$, both $SW$ and $IW$ become dominated by the globally 
highest-weight edges in the system. Specifically, 
$IW(\gamma) \sim |S_{\min}| \cdot e^{-\gamma c_{\min}}$ where 
$c_{\min} = \min_{i,j} c_{ij}$ and 
$S_{\min} = \{i : c^*_i = c_{\min}\}$. When the capacity vector 
$\hat{k}$ permits the matching to include these globally maximum-weight edges, 
$SW(\gamma)$ is governed by the same dominant exponential and the ratio 
recovers. This asymptotic behavior is formalized in 
Proposition~\ref{prop:gamma_convergence}.

\begin{proposition}
\label{prop:gamma_convergence}
Let $c_{\min} = \min_{i,j} c_{ij}$ and 
$S_{\min} = \{i : c^*_i = c_{\min}\}$, and define 
$J^*_i = \{j : c_{ij} = c^*_i\}$. Then:
\begin{equation}
    \lim_{\gamma \to \infty} \frac{\text{SW}(\gamma)}{\text{IW}(\gamma)} 
    = \frac{\max_{z \in \mathcal{Z}} 
    \sum_{i \in S_{\min}} \sum_{j \in J^*_i} z_{ij}}
    {|S_{\min}|}
    \label{eq:convergence}
\end{equation}
The limit equals 1 when $\hat{k}$ permits all seekers in $S_{\min}$ 
to access their best choice provider, and is less than 1 when capacity 
constraints prevent this. The limit depends on whether the initial capacity vector $\hat{k}$ can accommodate such minimum-cost assignments for seekers in $S_{min}$, not on the absolute magnitudes of the costs. The proof is provided in~\Cref{app:proof_convergence}.
\end{proposition}

%
The U-shaped curve reveals that $SW/IW$ is not monotone in $\gamma$, 
and that neither extreme ($\gamma = 0$ or $\gamma \to \infty$) is 
appropriate for practical use. The elbow region identifies the most 
informative operating range: $\gamma$ should be large enough that 
cost differences between providers are meaningfully reflected in the 
matching weights, but small enough that near-optimal alternatives 
remain distinguishable for seekers who cannot access their top-choice 
provider due to capacity constraints. When $\gamma$ is very small, the matching is entirely insensitive to costs; and when it is very large, the matching treats all non-top-choice alternatives  as identical, discarding useful information precisely when capacity 
constraints force displacement. In practice, we recommend identifying 
the elbow of the $SW/IW$ curve, the range where the curve transitions 
from steep decay to recovery, as the operating region for $\gamma$. 
In our large-scale experiments this corresponds to $\gamma \in 
[50, 150]$, and we therefore use $\gamma = 100$ as our operating 
value. Importantly, the relative ordering of the three optimization 
equations and all qualitative findings remain stable across this 
range, confirming that our conclusions are robust to the precise 
choice of $\gamma$.

%% file: sections/two-moons-results.tex
\begin{figure*}[t]
    \centering
    \Description{Experiment on 3-layer framework}
    \begin{subfigure}[b]{0.36\textwidth}         
        \centering
        \includegraphics[width=.9\linewidth]{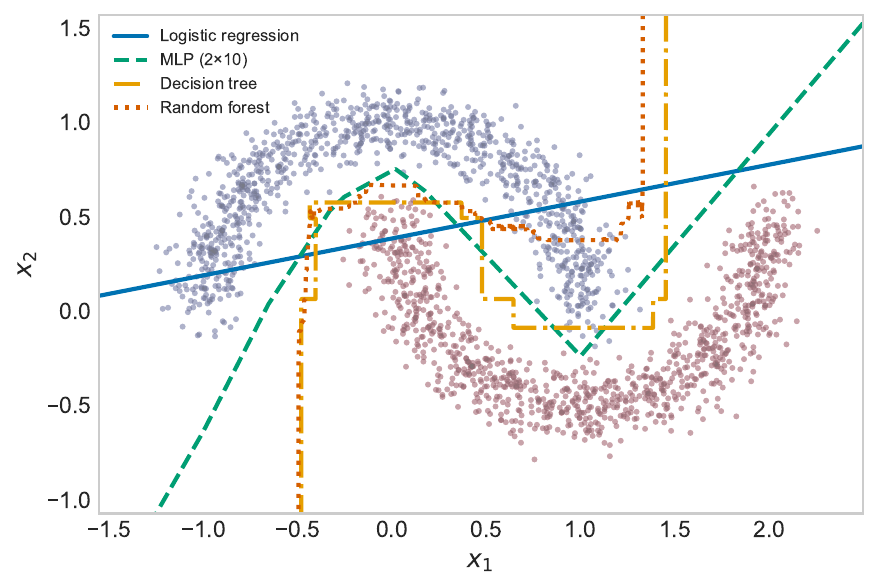}
        \caption{Two-Moon dataset overlaid with the decision boundaries of the four provider models, underscoring that seekers face genuinely different models.}
        \label{fig:exp-A}
    \end{subfigure}
    \hfill
    \begin{subfigure}[b]{0.24\textwidth} 
        \centering
        \resizebox{\linewidth}{!}{%
        $\mathbf W = \begin{bmatrix}
        0.548 & 0.425 & \colorbox{lightgray}{0.611} & 0.594 \\
        0.667 & 0.413 & 0.695 & \colorbox{lightgray}{0.703} \\
        0.539 & 0.488 & 0.521 & \colorbox{lightgray}{0.949} \\
        0.488 & \colorbox{lightgray}{0.687} & 0.666 & 0.484 \\
        0.773 & 0.354 & 0.793 & \colorbox{lightgray}{0.896} \\
        0.576 & 0.582 & 0.557 & \colorbox{lightgray}{0.834} \\
        0.558 & \colorbox{lightgray}{0.765} & 0.545 & 0.626 \\
        0.417 & 0.557 & \colorbox{lightgray}{0.558} & 0.415 \\
        \end{bmatrix}$%
        }
        \caption{Illustration of \Cref{alg:optimal_capacity} on the MACE-generated weight matrix. For each seeker (row), the largest edge weight \( w^*_i \) is highlighted; its column index \( j^*_i \) designates the preferred provider.}
        \label{fig:exp-B}
    \end{subfigure}
    \hfill
    \begin{subfigure}[b]{0.36\textwidth}         
        \centering
        \includegraphics[width=\linewidth]{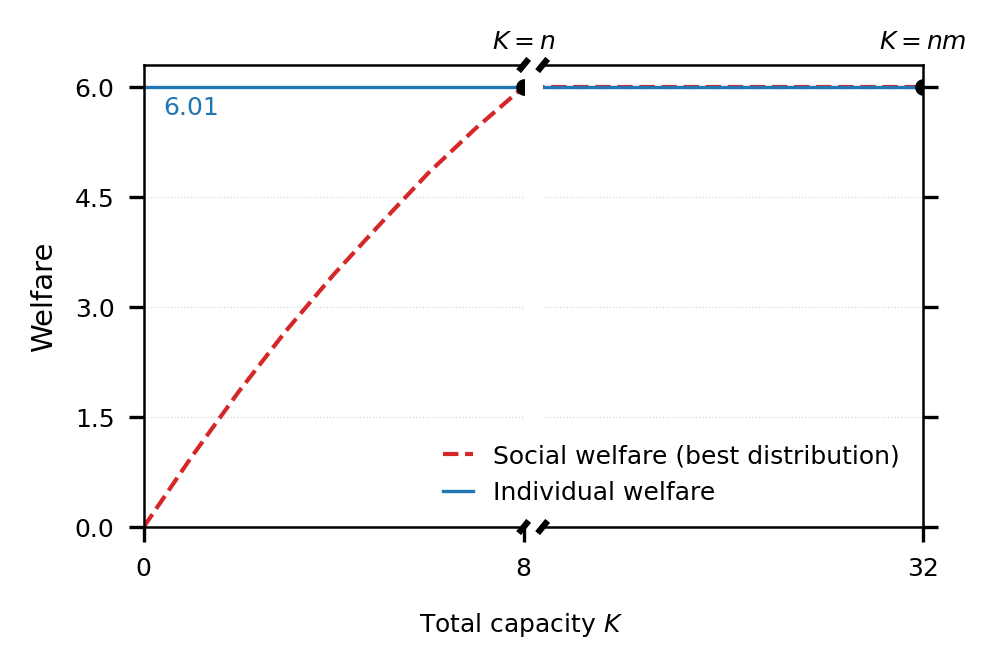}
        \caption{Social welfare attained by Algorithm 1 (red dashed) as the total capacity \(K\) grows from 0 to \(n*m\), against the individual-welfare upper-bound.}
        \label{fig:exp-C}
    \end{subfigure}

    \vspace{0.8em}                               

    \begin{subfigure}[b]{0.32\textwidth}         
        \centering
        \includegraphics[width=\linewidth]{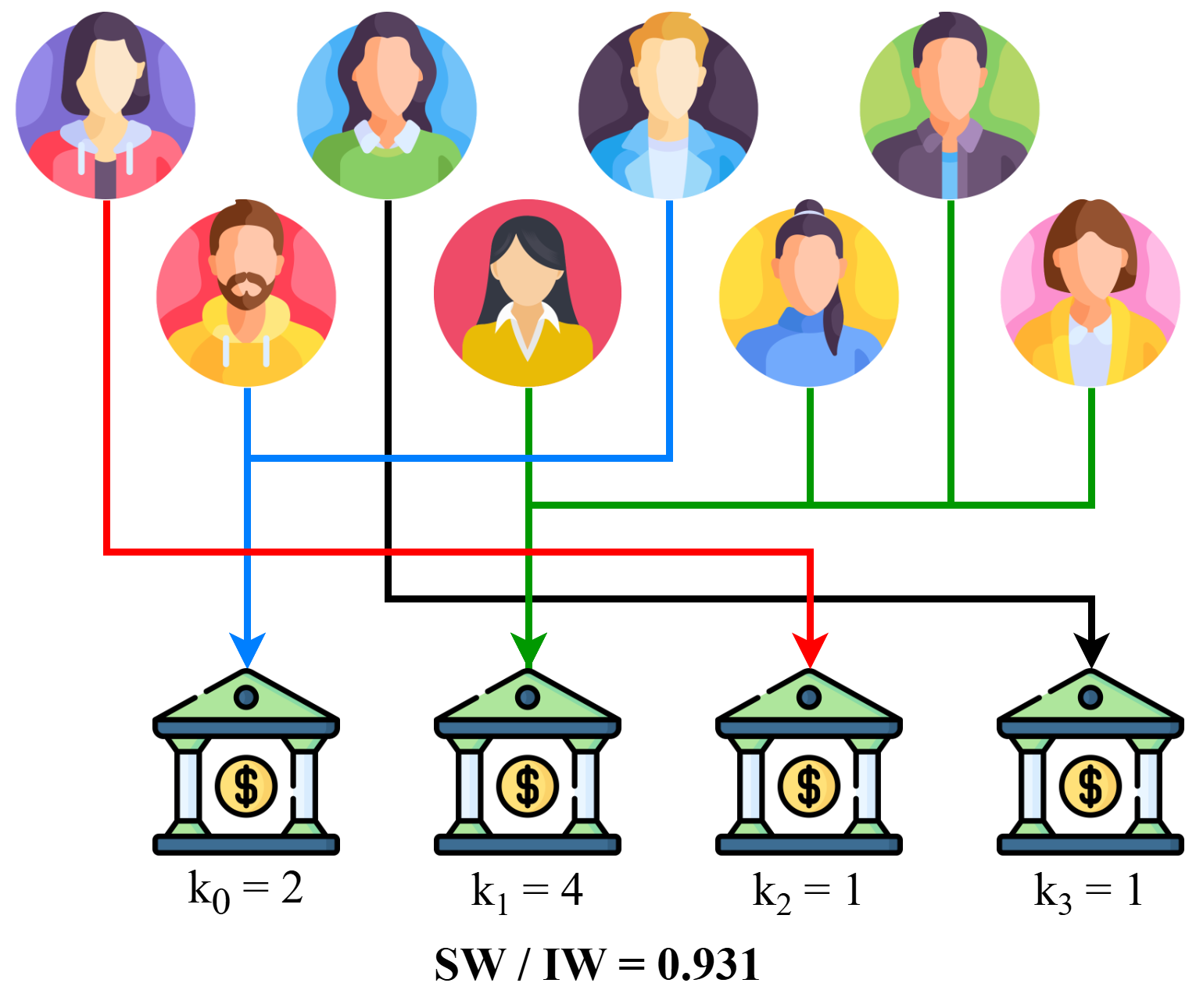}
        \caption{Optimal seeker-to-provider assignments obtained by solving \Cref{equ:first-optimization} under the initial capacity vector.}
        \label{fig:exp-D}
    \end{subfigure}
    \hfill
    \begin{subfigure}[b]{0.32\textwidth}         
        \centering
        \includegraphics[width=\linewidth]{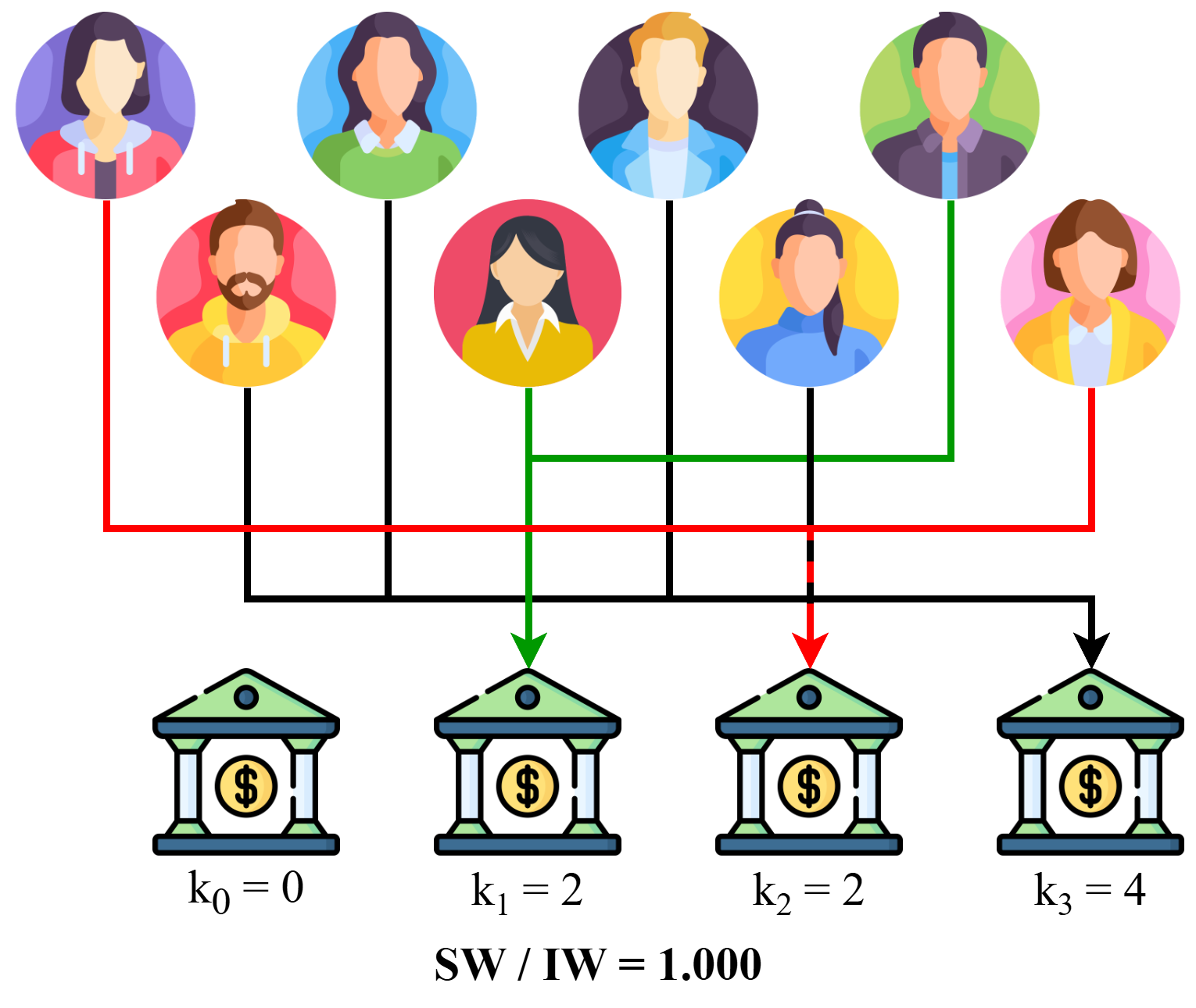}
        \caption{Provider capacities are distributed according to the capacity vector founded by \Cref{equ:second_optimization}, eliminating the welfare gap.}
        \label{fig:exp-E}
    \end{subfigure}
    \hfill
    \begin{subfigure}[b]{0.32\textwidth}         
        \centering
        \includegraphics[width=\linewidth]{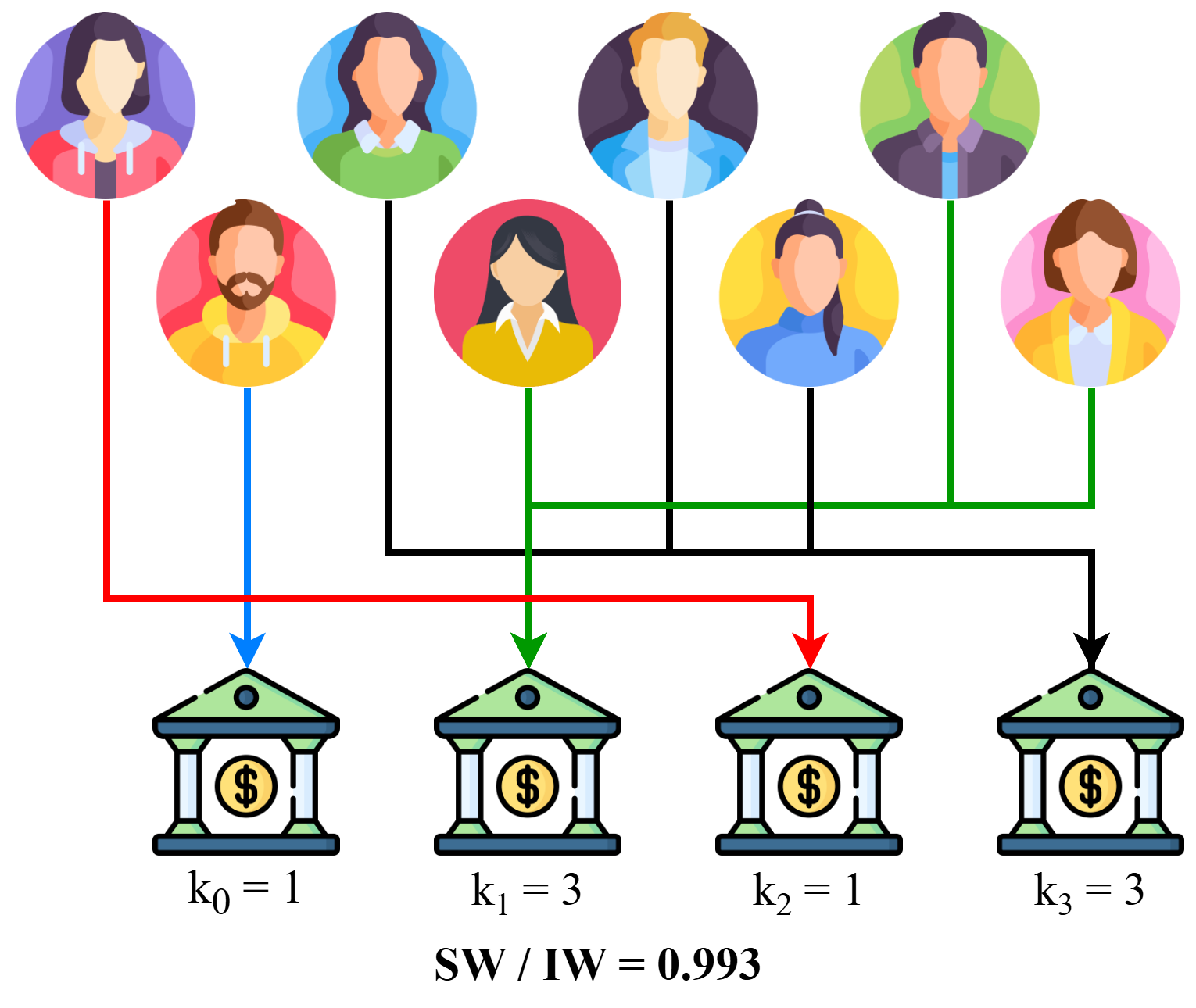}
        \caption{Matching produced by \Cref{equ:third_optimization}, balancing social-welfare against the cost of deviating from the initial capacities.}
        \label{fig:exp-F}
    \end{subfigure}

    \caption{Comprehensive illustration of our proposed framework on the Two Moon dataset and using the MACE method. (a) model decision boundaries, (b) weight matrix showing seeker-provider preferences and individually optimal matches, (c) welfare curve showing how total capacity impacts social welfare and its gap from the individual-welfare upper bound, (d) and the step-by-step optimization of provider capacities, (e) transitioning from initial assignments, to welfare-maximizing allocations, and finally (f) balancing welfare gains with capacity adjustment costs .}
    \label{fig:exp-sixpanel}
\end{figure*}

%% file: sections/Discussion.tex
\section{Discussion}
Our experiments show that welfare losses arise immediately in multi-agent settings when seekers apply independently under fixed, random capacities, even in low-dimensional feature spaces (e.g., the Two-Moon dataset). 
However, these losses can be substantially reduced with minimal system-level intervention.
When total system capacity matches the number of seekers, a single pass of \Cref{alg:optimal_capacity} restores the full social welfare.
This suggests that most of the efficiency gap stems not from resource scarcity but from poor allocation.
Moreover, introducing a modest penalty for deviating from the initial capacity distribution (\Cref{equ:third_optimization}) had little impact on overall performance. 
This indicates that near-optimal outcomes are achievable through small, targeted adjustments, supporting practical deployment.

The results also highlight the importance of model diversity. 
Providers whose classifiers align well with the underlying feature distribution of a particular sub-population accrue many high-weight edges. 
Directing additional capacity to such specialists both increases overall welfare and improves individual actionability. 
In contrast, giving every provider the same capacity can waste resources on providers that do not add much value.

%

While our results demonstrate the effectiveness of the proposed framework across datasets and settings, it is important to note that several design choices can influence the specific quantitative outcomes in figures and tables. 
These include the choice of the edge-weight transformation function, the scaling parameter \(\gamma\), and the penalty coefficients \(\beta_j\) used in cost-aware optimization. 
However, the central insight remains robust; there are fundamental differences between isolated one-to-one settings and resource-constrained many-to-many environments that need to be accounted for in system-level optimization. 
Looking forward, this framework can be enriched by considering game-theoretic extensions in which providers act strategically. 
Such formulations could capture competitive or decentralized recourse environments, opening new directions for modeling strategic behavior and fairness in multi-agent recourse systems.

Finally, our inequality-averse formulation reframes algorithmic recourse allocation as a system-design problem rather than a purely efficiency-driven optimization task. 
The choice of social welfare objective implicitly encodes assumptions about whose outcomes should be prioritized when capacity is scarce. 
In practice, this choice may be constrained by organizational risk tolerance or domain-specific fairness norms, suggesting that recourse systems should expose their welfare objectives as design parameters rather than fixed technical choices.

%% file: sections/Conclusion.tex
\section{Conclusion}
We have introduced a many-to-many view of algorithmic recourse in which multiple seekers obtain recommendations from multiple decision-making models whose resources are limited, and demonstrated that individually optimal recourse recommendations may no longer remain valid when real-world capacity constraints and stakeholder interactions come into play.
Further, we quantified the welfare gap between the socially optimal solution, computed by a central planner, and the individually optimal outcome, where each seeker selects the provider offering the lowest recourse cost, without coordination.  
Experiments demonstrated that explicit capacity management improves social welfare to the near-optimal upper bound while modifying minor capacity units.  
We have demonstrated how recourse allocation systems can incorporate inequality-averse objectives that explicitly prioritize individuals with the poorest outcomes. 
This extension highlights that recourse is not solely a technical optimization problem, but also a value-laden system design choice in which efficiency and fairness must be carefully balanced when resources are limited.

Beyond the static setting studied here, several adjustments deserve closer attention. 
On the seeker side, adding or removing applicants or allowing limited edits to their feature vectors directly reshapes the social welfare landscape. 
On the provider side, changing the number of providers or altering classifier parameters transforms the attainable optimum and the route toward it. 
Future studies can link these adjustment options to policy goals like ensuring diversity, fairness, or profit.
Also, real recourse ecosystems are not one-shot games; seekers reapply, models retrain, and resources fluctuate. 
Embedding the matching layer inside an online setting would let the system adapt capacities and decision boundaries in real time, continuously shrinking the welfare gap as new data arrive. 
Overall, the proposed framework bridges the space between individual prescriptions and system-wide outcomes, offers a tractable path toward higher social welfare, and opens several promising directions for adaptive and fairness-aware extensions.


%% file: sections/Appendix.tex
\section{Proof of Proposition~\ref{prop:gamma_convergence}}
\label{app:proof_convergence}

We prove that:
\begin{equation*}
    \lim_{\gamma \to \infty} \frac{\text{SW}(\gamma)}{\text{IW}(\gamma)} 
    = \frac{\max_{z \in \mathcal{Z}} 
    \sum_{i \in S_{\min}} \sum_{j \in J^*_i} z_{ij}}
    {|S_{\min}|}
\end{equation*}

\paragraph{Notation.}
Let $c_{\min} = \min_{i,j} c_{ij}$ denote the globally smallest 
recourse cost across all seeker-provider pairs. Let 
$c^*_i = \min_j c_{ij}$ denote the best achievable cost for seeker $i$, 
and define the set of seekers with globally minimal best cost:
\begin{equation*}
    S_{\min} = \{i : c^*_i = c_{\min}\}.
\end{equation*}
For each seeker $i$, define the set of minimum-cost providers:
\begin{equation*}
    J^*_i = \{j : c_{ij} = c^*_i\}.
\end{equation*}
The individual welfare and social welfare are:
\begin{equation*}
    \text{IW}(\gamma) = \sum_{i=1}^{n} e^{-\gamma c^*_i}, \qquad
    \text{SW}(\gamma) = \max_{z \in \mathcal{Z}} \sum_{i=1}^{n} 
    \sum_{j=1}^{m} e^{-\gamma c_{ij}} z_{ij},
\end{equation*}
where $\mathcal{Z}$ is the finite set of feasible binary matchings 
under the capacity and matching constraints of Equation~(\ref{eq:matching}). 
Since $\mathcal{Z}$ is finite, the maximum in $\text{SW}(\gamma)$ is always 
attained.

\paragraph{Step 1: Factor out the dominant exponential term.}
Factor $e^{-\gamma c_{\min}}$ from both numerator and denominator:
\begin{equation*}
    \frac{\text{SW}(\gamma)}{\text{IW}(\gamma)} = 
    \frac{\max_{z \in \mathcal{Z}} \sum_{i=1}^{n} \sum_{j=1}^{m} 
    e^{-\gamma(c_{ij} - c_{\min})} z_{ij}}
    {\sum_{i=1}^{n} e^{-\gamma(c^*_i - c_{\min})}}.
\end{equation*}
Since $c_{ij} \geq c^*_i \geq c_{\min}$ for all $i, j$, every 
exponent in both numerator and denominator is non-negative, so all 
terms lie in $[0, 1]$ for $\gamma > 0$.

\paragraph{Step 2: Limit of the denominator.}
For each seeker $i$:
\begin{equation*}
    e^{-\gamma(c^*_i - c_{\min})} \xrightarrow{\gamma \to \infty} 
    \begin{cases} 
    1 & \text{if } i \in S_{\min}, \\ 
    0 & \text{if } i \notin S_{\min}.
    \end{cases}
\end{equation*}
since $c^*_i - c_{\min} = 0$ for $i \in S_{\min}$ and 
$c^*_i - c_{\min} > 0$ otherwise. By dominated convergence (the sum is finite):
\begin{equation*}
    \sum_{i=1}^{n} e^{-\gamma(c^*_i - c_{\min})} 
    \xrightarrow{\gamma \to \infty} |S_{\min}|.
\end{equation*}

\paragraph{Step 3: Limit of each term in the numerator.}
For any fixed feasible matching $z \in \mathcal{Z}$ and any 
seeker-provider pair $(i,j)$:
\begin{equation*}
    e^{-\gamma(c_{ij} - c_{\min})} \xrightarrow{\gamma \to \infty} 
    \begin{cases} 
    1 & \text{if } c_{ij} = c_{\min}, \\ 
    0 & \text{if } c_{ij} > c_{\min}.
    \end{cases}
\end{equation*}
Now $c_{ij} = c_{\min}$ is possible only when $i \in S_{\min}$ and 
$j \in J^*_i$, since for $i \in S_{\min}$ we have $c^*_i = c_{\min}$.
Therefore, for any fixed $z \in \mathcal{Z}$, since each seeker is matched 
to at most one provider ($\sum_j z_{ij} \leq 1$):
\begin{equation*}
    \sum_{i=1}^{n} \sum_{j=1}^{m} e^{-\gamma(c_{ij} - c_{\min})} 
    z_{ij} \xrightarrow{\gamma \to \infty} 
    \sum_{i \in S_{\min}} \sum_{j \in J^*_i} z_{ij},
\end{equation*}
which counts the number of seekers in $S_{\min}$ matched to one of 
their minimum-cost providers under $z$.

\paragraph{Step 4: Pointwise limit of $F(z, \gamma)$.}
For each fixed $z \in \mathcal{Z}$, define:
\begin{equation*}
    F(z, \gamma) = \frac{\sum_{i=1}^{n} \sum_{j=1}^{m} 
    e^{-\gamma(c_{ij} - c_{\min})} z_{ij}}
    {\sum_{i=1}^{n} e^{-\gamma(c^*_i - c_{\min})}}.
\end{equation*}
By Steps 2 and 3, both numerator and denominator converge as 
$\gamma \to \infty$, and the denominator converges to $|S_{\min}| > 0$. 
Therefore:
\begin{equation*}
    F(z, \gamma) \xrightarrow{\gamma \to \infty} 
    \frac{\sum_{i \in S_{\min}} \sum_{j \in J^*_i} z_{ij}}{|S_{\min}|}.
\end{equation*}

\paragraph{Step 5: Exchange limit and maximum.}
Since $\mathcal{Z}$ is a \textit{finite} set, we can exchange the 
limit and the maximum:
\begin{equation*}
    \lim_{\gamma \to \infty} \frac{\text{SW}(\gamma)}{\text{IW}(\gamma)}
    = \lim_{\gamma \to \infty} \max_{z \in \mathcal{Z}} F(z, \gamma)
    = \max_{z \in \mathcal{Z}} \lim_{\gamma \to \infty} F(z, \gamma)
    = \frac{\max_{z \in \mathcal{Z}} 
    \sum_{i \in S_{\min}} \sum_{j \in J^*_i} z_{ij}}{|S_{\min}|}.
\end{equation*}

The right-hand side is the maximum fraction of seekers in $S_{\min}$ 
who can simultaneously be matched to one of their minimum-cost providers 
under the fixed capacity vector $\hat{k}$. It is determined by $\hat{k}$ 
and the structure of the minimum-cost edges $\{J^*_i\}_{i \in S_{\min}}$, 
and is independent of non-minimal cost magnitudes. $\hfill\square$

\section{Provider Classification Performance}
\label{app:provider_perf}
 
\Cref{tab:provider_performance} reports accuracy, precision, and recall
for all 15 providers used in the large-scale Credit and COMPAS experiments,
evaluated on the held-out test set using each provider's tier-specific
classification threshold.
The table confirms heterogeneity in predictive behavior across providers,
particularly in recall, which reflects the varying rates at which providers
reject applicants and consequently determines the difficulty of obtaining
recourse from each provider.
 
\begin{table}[h]
  \centering
  \caption{Classification performance of the 15 providers on the held-out test
           set (large-scale DiCE setting). Predictions use each provider's
           tier-specific threshold.}
  \label{tab:provider_performance}
  \small
  \begin{tabular}{llccc}
    \toprule
    \textbf{Provider} & \textbf{Accuracy} & \textbf{Precision} & \textbf{Recall} \\
    \midrule
    LR\_1 & 0.733 & 0.883 & 0.758 \\
    Tree\_2 & 0.703 & 0.893 & 0.703 \\
    RF\_3 & 0.699 & 0.891 & 0.698 \\
    MLP\_4 & 0.669 & 0.899 & 0.648 \\
    LR\_5 & 0.709 & 0.887 & 0.718 \\
    Tree\_6 & 0.703 & 0.893 & 0.703 \\
    RF\_7 & 0.652 & 0.900 & 0.622 \\
    MLP\_8 & 0.550 & 0.917 & 0.465 \\
    LR\_9 & 0.419 & 0.907 & 0.283 \\
    Tree\_10 & 0.221 & 0.000 & 0.000 \\
    RF\_11 & 0.755 & 0.881 & 0.793 \\
    MLP\_12 & 0.735 & 0.877 & 0.767 \\
    LR\_13 & 0.733 & 0.883 & 0.758 \\
    Tree\_14 & 0.796 & 0.851 & 0.895 \\
    RF\_15 & 0.745 & 0.874 & 0.785 \\
    \bottomrule
  \end{tabular}
\end{table}

\section{Additional Experiments}
\label{app:extra-exp}
\begin{figure*}[h]
    \centering
    \begin{minipage}{0.8\textwidth}
        \centering
        \begin{subfigure}[b]{0.32\textwidth}
            \centering
            \resizebox{\linewidth}{!}{%
            $\mathbf W = \begin{bmatrix}
            0.256 & 0.104 & 0.256 & \colorbox{lightgray}{0.353} \\
            0.383 & 0.096 & 0.366 & \colorbox{lightgray}{0.441} \\
            0.179 & 0.148 & 0.129 & \colorbox{lightgray}{0.902} \\
            0.415 & 0.064 & 0.432 & \colorbox{lightgray}{0.800} \\
            0.187 & 0.237 & 0.168 & \colorbox{lightgray}{0.694} \\
            0.299 & \colorbox{lightgray}{0.499} & 0.178 & 0.393 \\
            0.165 & \colorbox{lightgray}{0.214} & 0.198 & 0.171 \\
            0.211 & 0.021 & 0.230 & \colorbox{lightgray}{0.251} \\
            0.509 & 0.049 & 0.535 & \colorbox{lightgray}{0.727} \\
            0.368 & \colorbox{lightgray}{0.830} & 0.314 & 0.395
            \end{bmatrix}$%
            }
            \caption{Illustration of \Cref{alg:optimal_capacity} on the MACE-generated weight matrix.}
            \label{fig:l1-b}
        \end{subfigure}
        \hspace{2em}
        \begin{subfigure}[b]{0.45\textwidth}
            \centering
            \includegraphics[width=\linewidth]{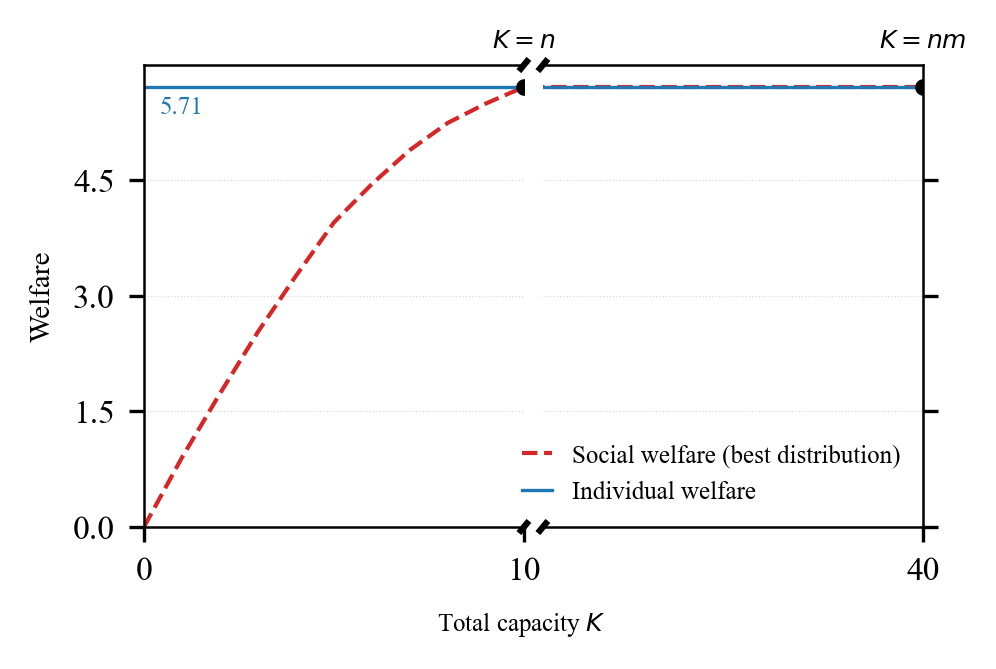}
            \caption{Social welfare and individual-welfare plotted as the total capacity K grows from 0 to \(n*m\).}
            \label{fig:l1-c}
        \end{subfigure}
    \end{minipage}

    \vspace{0.8em}

    \begin{subfigure}[b]{0.32\textwidth}
        \centering
        \includegraphics[width=\linewidth]{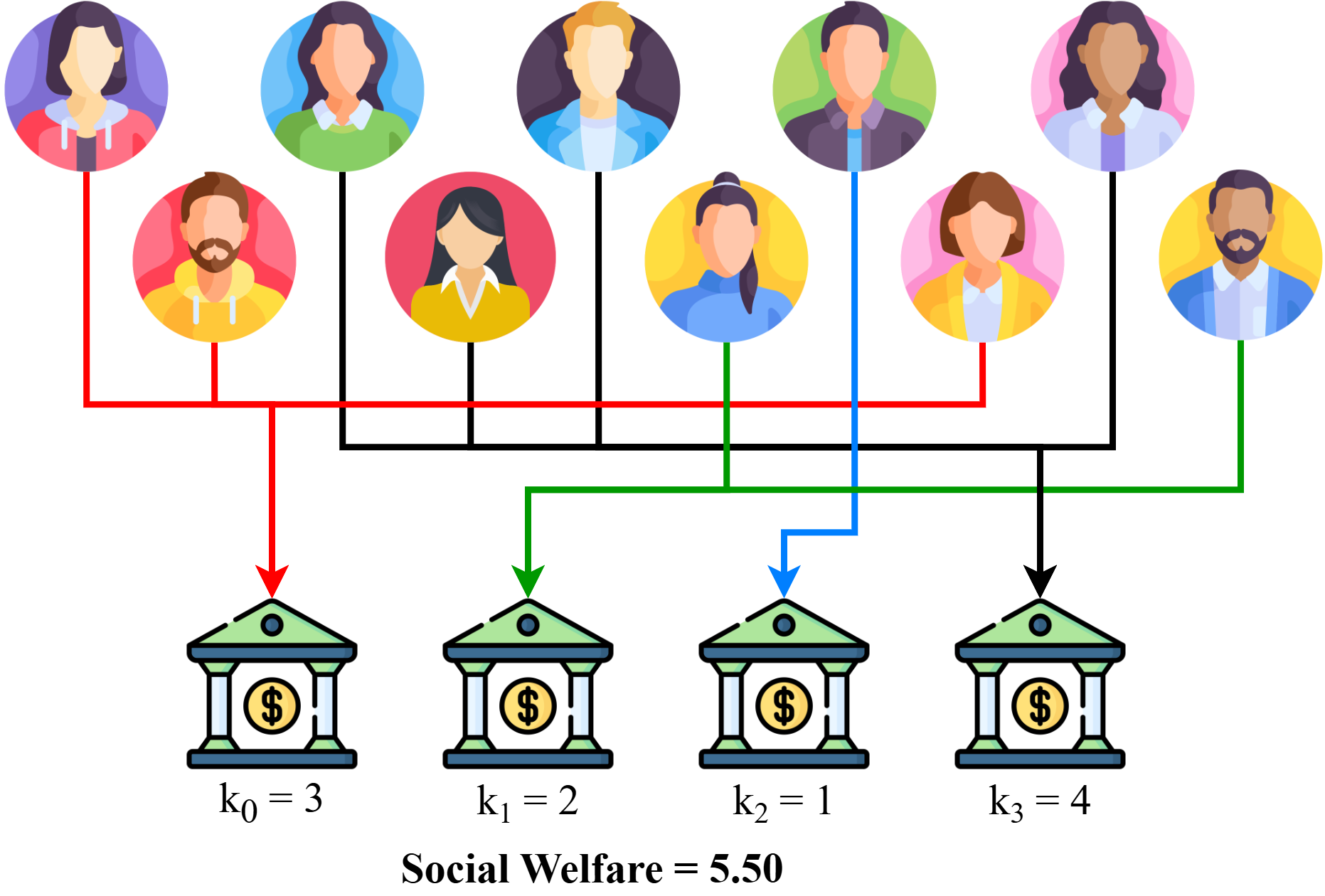}
        \caption{Optimal seeker-to-provider assignments obtained by solving \Cref{equ:first-optimization} under the initial capacity vector.}
        \label{fig:l1-d}
    \end{subfigure}
    \hfill
    \begin{subfigure}[b]{0.3\textwidth}
        \centering
        \includegraphics[width=\linewidth]{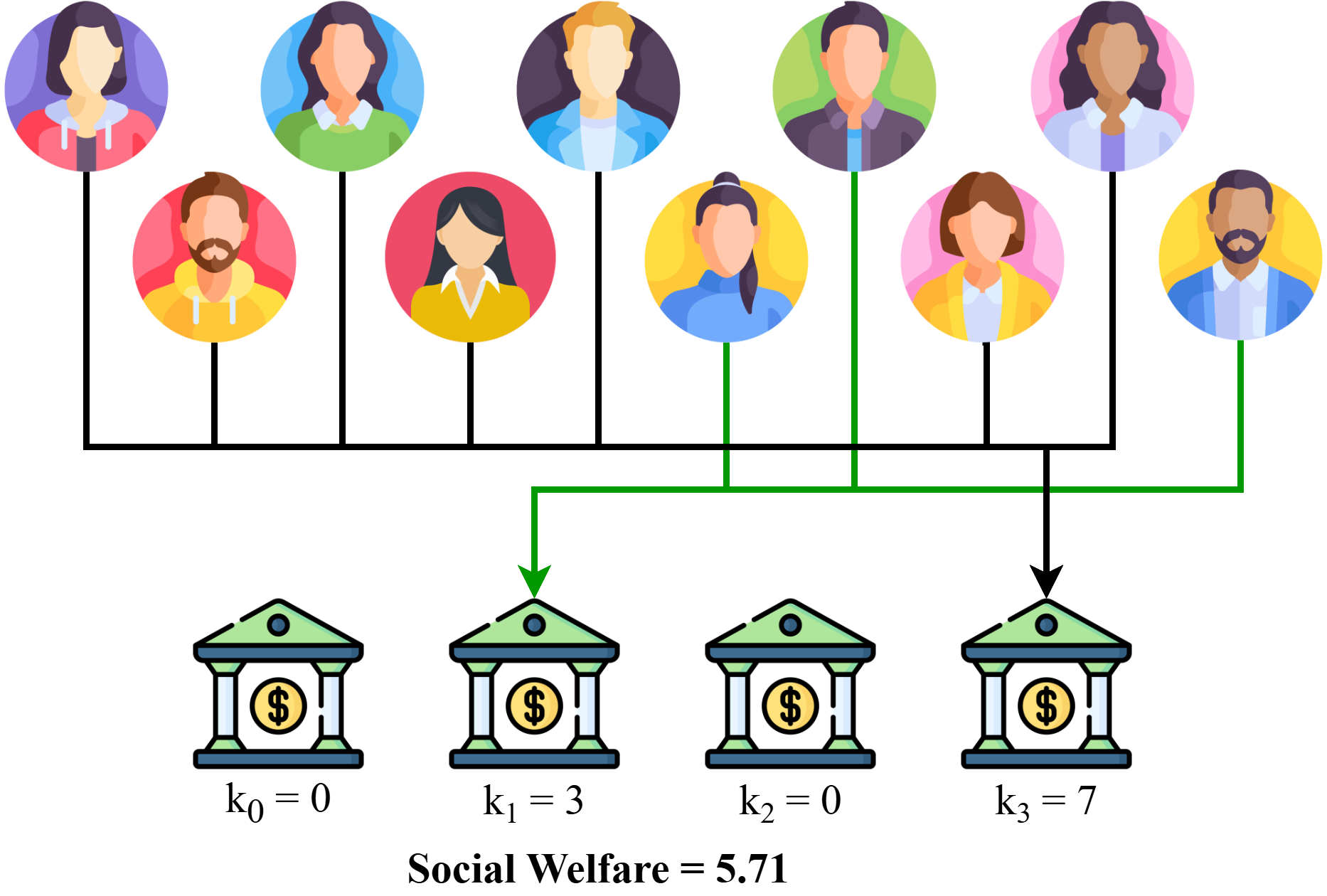}
        \caption{Provider capacities distributed via~\Cref{equ:second_optimization}, eliminating the welfare gap.}
        \label{fig:l1-e}
    \end{subfigure}
    \hfill
    \begin{subfigure}[b]{0.3\textwidth}
        \centering
        \includegraphics[width=\linewidth]{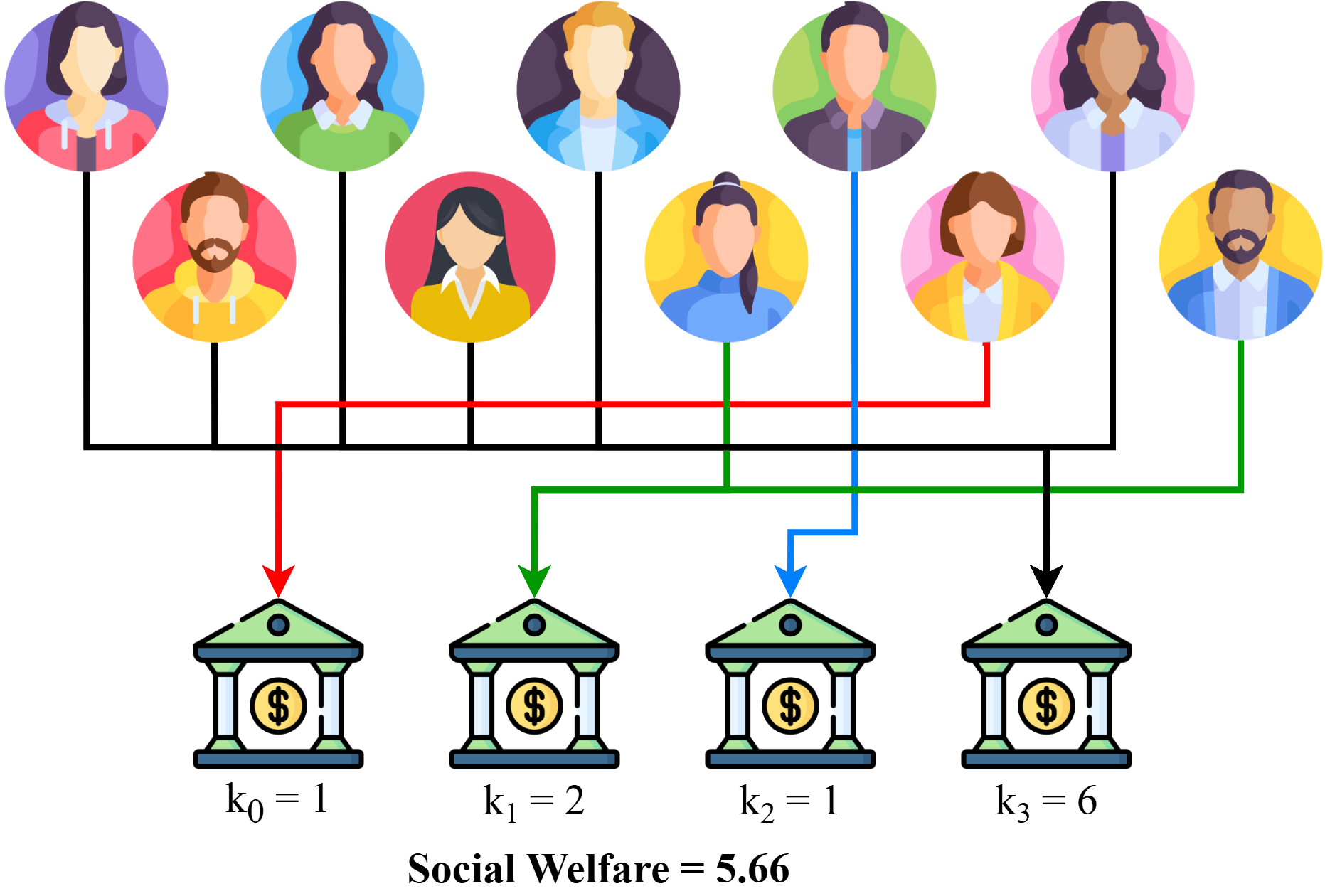}
        \caption{Final matching (\Cref{equ:third_optimization}) balancing welfare against capacity deviation costs.}
        \label{fig:l1-f}
    \end{subfigure}

    \caption{Illustration of our proposed framework: Weight matrix and individually optimal matches (a), Welfare curve showing the gap between social and individual optima. (c--e) Optimization process: transitioning from initial assignments (c) to welfare-maximizing allocations (d), and balancing welfare gains against capacity adjustment costs (e).}
    \label{fig:l1-results}
\end{figure*}

To test robustness to the distance function, we repeated all experiments with the $\ell_1$ norm instead of the $\ell_\infty$ norm. 
The datasets, provider models, and optimization procedure were unchanged, except for minor parameter adjustments. 
We kept the setup and $\gamma$ the same but increased $\beta$ to 0.05. 
As shown in \Cref{fig:l1-results}, the framework still achieves near-optimal welfare with minimal adjustments, and quantitative results for all datasets are summarized in \Cref{tab:l_1_results}.

\begin{table}[h]
\centering
\caption{Comparative results of the three optimization formulations. 
         Each cell shows social welfare (SW), capacity vector \(\mathbf{k}\), 
         and the percentage of the individual-welfare upper bound (IW) attained in each optimization problem.}
\label{tab:l_1_results}

\small
\setlength{\tabcolsep}{4pt}
\renewcommand{\arraystretch}{1.1}

    \begin{tabular}{l ccc ccc ccc}
    \toprule
      & \multicolumn{3}{c}{\textbf{Two-Moon}} 
      & \multicolumn{3}{c}{\textbf{Credit}} 
      & \multicolumn{3}{c}{\textbf{COMPAS}} \\
    \cmidrule(lr){2-4} \cmidrule(lr){5-7} \cmidrule(lr){8-10}
      & Capacity & SW & \%\,IW 
      & Capacity & SW & \%\,IW 
      & Capacity & SW & \%\,IW \\
    \midrule
    \Cref{equ:first-optimization} 
     & $(3,2,1,4)$ & 5.50 & 96.32\% 
     & $(3,2,6,1)$ & 7.49 & 87.04\% 
     & $(3,8,1,3)$ & 12.03 & 95.74\% \\
    \Cref{equ:second_optimization} 
     & $(0,3,0,7)$ & 5.71 & 100\% 
     & $(4,0,3,5)$ & 8.61 & 100\% 
     & $(11,1,3,0)$ & 12.57 & 100\% \\
    \Cref{equ:third_optimization} 
     & $(1,2,1,6)$ & 5.66 & 99.03\% 
     & $(3,0,6,3)$ & 8.44 & 98.03\% 
     & $(5,6,1,3)$ & 12.26 & 97.53\% \\
    \bottomrule
    \end{tabular}%
\end{table}